%% file: PaperForReview.tex
\documentclass[10pt,twocolumn,letterpaper]{article}

\usepackage[pagenumbers]{cvpr} % To force page numbers, e.g. for an arXiv version

\usepackage{graphicx}
\usepackage{amsmath}
\usepackage{amssymb}
\usepackage{amstext}
\usepackage{booktabs}
\usepackage{tablefootnote}
\usepackage{array,multirow}
\usepackage{fontawesome}
\usepackage{cite}
\usepackage{amsfonts}
\usepackage{algorithmic}
\usepackage{textcomp}
\usepackage{xcolor}
\usepackage{overpic}
\usepackage[style=base]{caption}
\usepackage{subcaption}
\makeatletter
\@namedef{ver@everyshi.sty}{}
\makeatother
\usepackage{tikz}
\usepackage{transparent}
\usetikzlibrary{shapes,patterns,shapes.geometric,arrows.meta,bending,positioning}
\usepackage{pgfplots}
\pgfplotsset{compat=newest}
\usepackage{pgf-pie}
\def\BibTeX{{\rm B\kern-.05em{\sc i\kern-.025em b}\kern-.08em
    T\kern-.1667em\lower.7ex\hbox{E}\kern-.125emX}}

\usepackage[normalem]{ulem}

\newcolumntype{C}{>{$}c<{$}}

\input{definitions.tex}

\usepackage[pagebackref,breaklinks,colorlinks]{hyperref}

\usepackage[capitalize]{cleveref}
\crefname{section}{Sec.}{Secs.}
\crefname{section}{Section}{Sections}
\crefname{table}{Table}{Tables}
\crefname{table}{Tab.}{Tabs.}

\def\cvprPaperID{10859} % *** Enter the CVPR Paper ID here
\def\confName{CVPR}
\def\confYear{2023}

\begin{document}

\title{Towards Rapid Prototyping and Comparability in Active Learning for Deep Object Detection}

\author{Tobias Riedlinger${}^{*1}$ \quad Marius Schubert${}^{*1}$ \quad Karsten Kahl${}^{1}$ \quad Hanno Gottschalk${}^{1}$ \quad Matthias Rottmann${}^{1,2}$\\
$^{1}$ School of Mathematics and Natural Sciences, IZMD, University of Wuppertal, Germany\\
$^{2}$ School of Computer and Communication Sciences, CVLab, EPFL, Switzerland \\
{\tt\small \{riedlinger@math., mschubert@, kkahl@math., hgottsch@, rottmann@math.\}uni-wuppertal.de}
% \and
% \and
% \and
% \and
}
\maketitle

%%%%%%%%% ABSTRACT
\begin{abstract}
    Active learning as a paradigm in deep learning is especially important in applications involving intricate perception tasks such as object detection where labels are difficult and expensive to acquire.
    Development of active learning methods in such fields is highly computationally expensive and time consuming which obstructs the progression of research and leads to a lack of comparability between methods.
    In this work, we propose and investigate a sandbox setup for rapid development and transparent evaluation of active learning in deep object detection.
    Our experiments with commonly used configurations of datasets and detection architectures found in the literature show that results obtained in our sandbox environment are representative of results on standard configurations.
    The total compute time to obtain results and assess the learning behavior can thereby be reduced by factors of up to 14 when comparing with Pascal VOC and up to 32 when comparing with BDD100k.
    This allows for testing and evaluating data acquisition and labeling strategies in under half a day and contributes to the transparency and development speed in the field of active learning for object detection.
\end{abstract}

\def\thefootnote{*}\footnotetext{Equal contribution}

%%%%%%%%% BODY TEXT
\section{Introduction} % 1.5 P
% Herleitung der Problemstellung (Active Learning as future technology?)
% Warum das wichtig ist / mit Beobachtungen / Rechenaufwand etc.?
% Eigene Contributions
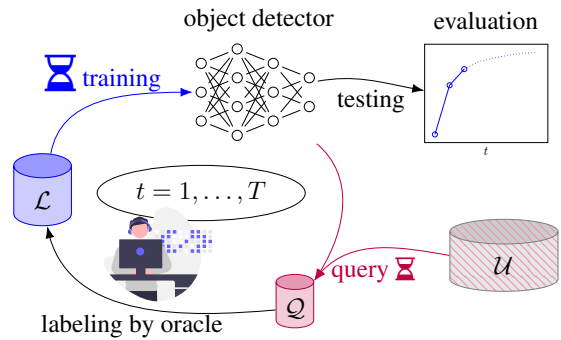
\begin{figure}
    \centering
    \resizebox{0.9\linewidth}{!}{
    \input{figures/cycle}
    }
    \caption{The generic pool-based AL cycle consisting of training on labeled data $\mathcal{L}$, querying informative data points $\mathcal{Q}$ out of a pool of unlabeled data $\mathcal{U}$ and annotation by a (human) oracle.
    % The evaluation phase assessing the effectiveness of an AL strategy can be performed in parallel with the query.
    In practice, training compute time is orders of magnitude larger than evaluating the AL strategy itself.
    }
    \label{fig: al cycle}
\end{figure}
\noindent
Deep learning requires large amounts of data, typically annotated by vast amounts of human labor \cite{zhan2022comparative,budd2021survey,li2006confidence}.
In particular in complex computer vision tasks such as object detection (OD), the amount of labor per image can lead to substantial costs for data labeling.
Therefore, it is desirable to avoid unnecessary labeling effort and to have a rather large variability of the database. 
\emph{Active learning} (AL)\cite{settles.tr09} is one of the key methodologies that aims at labeling the data that matters for learning. 
AL alternates model training and data labeling as illustrated in \cref{fig: al cycle}. 
At the core of each AL method is a query strategy that decides after each model training which unlabeled data to query for labeling. 
The computation cost of AL is in general at least an order of magnitude higher than ordinary model training and so is its development \cite{tsvigun2022towards,li2006confidence}, which comprises several AL experiments involving $T$ query steps with different parameters, ablation studies, etc. 
Hence, it is notoriously challenging to develop new AL methods for applications where model training itself is already computationally costly. 
In the field of OD, a number of works overcame this cumbersome hurdle \cite{yoo2019learning,brust2018active,roy2018deep,haussmann2020scalable,schmidt2020advanced,choi2021active,yuan2021multiple,elezi2021not,papadopoulos2017training,desai2019adaptive,subramanian2018one}. 
However, these works did so in very different settings which makes their comparison difficult. 
Hence, it is difficult to give any advice for practitioners, which AL method to choose. 
Besides that, AL with real-world data may suffer from other influencing factors, \eg, the quality of labels to which end fundamental research is conducted on AL in presence of label errors \cite{bouguelia2015identifying,bouguelia2018agreeing,younesian2020active,younesian2021qactor}.
These observations demand for a development environment that enables rapid prototyping, cutting down to the huge computational efforts of AL in object detection and fostering comparability of different AL methods.

\paragraph*{Contribution.} %\HG{HG: Auf a) folgt hier kein b)}
In this work, we propose a development environment that drastically cuts down the computational cost of developing AL methods. 
We do so by reducing the complexity of the deep object detectors as well as the complexity of the data base, establishing a non-trivial sandbox for deep OD.
To this end, we construct a) two datasets that generalize MNIST digits \cite{lecun1998gradient} and EMNIST letters \cite{cohen2017emnist} to the setting of OD by pasting colored and transformed samples into background images from MS-COCO \cite{lin2014microsoft} and b) a selection of suitable small-scale object detectors.
% \HG{Dass eine Tätigkeit 'psasting' mit einem Objekt 'object detector' accompanied wird, wirkt in meiner Vorstellung etwas kompliziert.}
We justify this step and underline its value for the field of AL in OD. To do so, we conduct several experiments which show that results on our datasets generalize to a similar degree to established but more complex datasets (Pascal VOC\cite{Everingham10} and BDD100k \cite{yu2020bdd100k}), as they generalize among each other. %\HG{HG 'being representative' ist ein nicht so präzises Verb. Bitte genauer.}
We also demonstrate that we reduce the computational effort %of both training and inference (required for evaluation and querying) 
of AL experiments by factors of up to $32$. %\HG{Sagen, worzu der computational effort dient.}
In addition, to establish further comparability of AL methods for OD, we propose an extensive evaluation protocol.
We summarize our contributions as follows:
\begin{itemize}
    % \item We show that active learning histories are strongly influenced by the type of architecture and dataset used, making setups difficult to compare. \HG{Ok, aber haben wir nicht gerade für das Gegenteil plädiert?}
    \item We propose a sandbox environment with two datasets, multiple networks, active learning queries and an extensive evaluation protocol. This setup allows for broader comparisons and detailed and transparent experiment tracking at lowered computational effort. We demonstrate this with extensive numerical results. % and improved development accuracy and time.} %\HG{Auch das widerspricht dem bullet oben.}
    \item We analyze the generalization capabilities of our sandbox and find that results obtained by our sandbox generalize well to Pascal VOC and BDD100k.
    \item We contribute to future AL development by providing an implementation of our pipeline in a flexible environment as well as an automated framework for evaluation and visualization of results. 
    % \HG{Bei ready-to-plot erwartet man nach dem Beginn des Bulletpoints Auswertungstools, es kommen aber Experimente.}
\end{itemize}

Code will be publicly available at \texttt{GitHub}. 
All selected configurations and hyperparameters in addition to evaluation files for the experiments conducted will be published in order to facilitate future comparisons with the implemented baselines.

The remainder of this work is structured as follows: 
\cref{sec: related work} contains a summary of the literature in fully-supervised AL for OD and how the present work relates to it.
In \cref{sec: sandbox environment} we introduce our sandbox setup for AL in OD, methods involved in our experiments and our proposed evaluation metrics.
\cref{sec: developing al} first introduces our experimental setup.
We investigate the degree to which AL in OD is comparable in \cref{sec: benchmark results} by studying different cases.
Afterwards, in \cref{sec: experimental representativity} we investigate rank correlations between AL on different datasets to measure the degree of similarity between the results.
In \cref{sec: compute time}, we show time measurements to estimate the speed-up provided by our sandbox environment.
Finally, we close with a summary of conclusions we draw from the empirical evidence in \cref{sec: conclusion}.

\section{Related Work} % 0.5 - 1 P
\label{sec: related work}
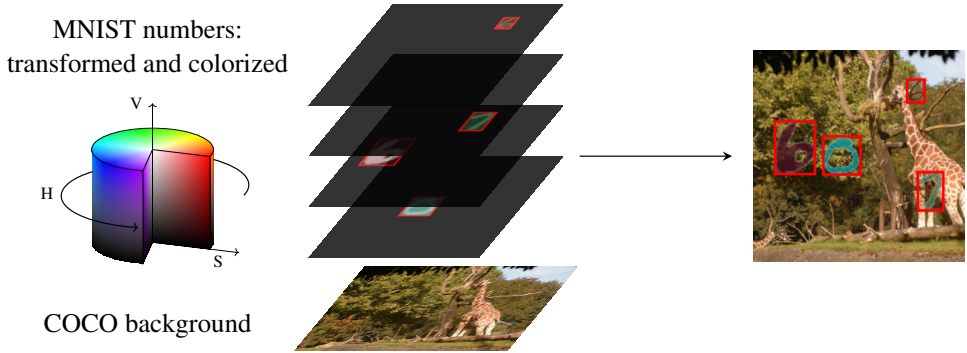
\begin{figure*}
    \centering
    \resizebox{0.75\textwidth}{!}{\input{figures/diagram}}
    \caption{Generation scheme of semi-synthetic object detection data from MNIST digits on a non-trivial background image from MS-COCO.}
    \label{fig: dataset generation}
\end{figure*}
\noindent
Numerous methods of AL have been, and still are, developed in the classification setting and largely fall into two main categories.
Uncertainty-based query strategies rely on the informativeness of a probabilistic model's %(or that of a potentially small ensemble) 
current prediction uncertainty\cite{lewis1994heterogeneous}.
%todo: Quelle für Entropysampling.
In probabilistic classification models, popular examples include the probability margin\cite{scheffer2001active}, ensemble entropy\cite{dagan1995committee}, or committee disagreement\cite{seung1992query}.
A different approach is taken by density-based query strategies, which aim at exploiting data diversity between already known data and prospective queries.
Typically, such methods are related to clustering algorithms\cite{nguyen2004active, xu2007incorporating, sener2017active} which can also be used in conjunction with uncertainty estimation methods. 
For a general overview of different AL approaches see e.g.\cite{settles.tr09}.

The task of OD is more complex than image classification in that the background needs to be distinguished against foreground instances and each foreground instance needs to be assigned its own localization and classification.
Not only does this require the choice of additional hyperparameters for making predictions, but this also entails more complex and expensive labels.
Therefore, AL plays also a large role in OD.
% \HG{Hier kommt der wesentliche Unterschied zu Klassifikation, nämlich die Rolle des Hintergrundes, nicht gut raus.}
Yoo and Kweon\cite{yoo2019learning} present a task-agnostic method where a loss prediction module estimates a loss for every potentially queried image and selects the images for which the largest loss is expected.
% A task-agnostic method is presented in~\cite{yoo2019learning}, a loss prediction module estimates a loss for every potentially queried image and selects the images for which the largest loss is expected.
Brust \etal\cite{brust2018active} estimate prediction-wise uncertainty by the probability margin and aggregate to image uncertainty by the summation, averaging or taking the maximum.
% In~\cite{brust2018active} prediction-wise uncertainty is estimated by the probability margin and image-wise uncertainty is then estimated by the sum, the average or the maximum of the prediction-wise uncertainties.
For a black-box approach, Roy \etal\cite{roy2018deep} follow the same idea, but with using the classification entropy as prediction uncertainty. 
Moreover, they propose a white-box approach which is inspired by query-by-committee, making use of predictions at different scales in the object detector.
% Whereas, the proposed white-box approach was inspired by query by committee~\cite{melville2004diverse} and estimates a instance-wise uncertainty based on maximal classification probability margins of significant overlapping boxes before Non Maximum Suppression (NMS).
In general, an ensemble of OD heads, trained independently on the same set of labeled images, can be used to estimate prediction-wise classification uncertainty, \ie mutual information based on entropies~\cite{haussmann2020scalable}, or a combination of localization and classification uncertainty due to region of interest similarities~\cite{schmidt2020advanced}.
But in particular, as training a variety of detector heads in each AL step is very costly, committee/ensemble query methods tend to be approximated by Monte Carlo (MC) Dropout~\cite{gal2016dropout, gal2017deep}.
Some methods yield to custom object detectors, \ie a Gaussian mixture model based object detector with an adjusted loss \cite{choi2021active} or they intervene at least in the training pipeline, \eg by suppressing noisy instances~\cite{yuan2021multiple}.
Another class of approaches is based on semi-supervised learning, \ie incorporating either pseudo-labels for some representative easy samples to prevent distribution drift~\cite{elezi2021not}.
Alternatively used ``weak'' labels consist only of the center coordinates~\cite{papadopoulos2017training,desai2019adaptive,subramanian2018one}.
In this paper we compare uncertainty-based methods with each other that are exclusively based on fully supervised training \cite{brust2018active,roy2018deep,haussmann2020scalable,choi2021active}.
% \TR{Statt dieser Liste (oder im Kontext) noch herausstellen, was die Novelty unserer Arbeit ausmacht und was genau der Zusammenhang (``Related'' Work) ist. Wie steht unsere Arbeit der Literatur entgegen?}
These works are difficult to compare since the datasets, network architectures, frameworks, initial dataset sizes, query sizes and hyperparameters for training and inference heavily differ from each other.
Unlike the works mentioned, we aim at putting the AL task itself on equal footing between different settings to improve development speed and evaluation transparency.
In our work, we compare some of the above-mentioned methods to each other based on the same configurations for frequently used datasets and network architectures.
Comparative investigations of this kind has escaped previous research in the field.
% This results in comparability of methods across the full Al cycle, analyzed with more robust metrics compared to the metrics used in the literature. This is indispensable for the development of new acquisition strategies that can be immediately compared with established baselines. In addition, our sandbox environment drastically reduces the development times that as soon as new queries are implemented the whole AL cycle for this can be evaluated within half a day. }
%\commentMS{\sout{In \cite{brust2018active} the Pascal VOC training and validation data are each split up in two disjoint sets to learn new classes. The underlying network is a YOLO architecture \cite{redmon2018yolov3} and 5 query steps are made, each of which is 50 samples large. The combination of the VOC2007 and VOC2012 trainval sets is used for training in \cite{roy2018deep}. The initial training set contains 10\% randomly selected images of this set and a query size of 5\% is used, up to a total volume of 45\% training images. A single shot multibox detector is used for training the model\cite{liu2016ssd}. The experiments in \cite{haussmann2020scalable} are based on an internal data set comprising 847,000 training images. The initial training happens on 100,000 images and 200,000 images are queried in each step until all images are involved in the training process. Each of the members of the ensemble is a one-stage object detector based on a UNet-backbone.}}

% \HG{HG: Kommentiert bis hier}
\section{A Sandbox Environment with Datasets, Models and Evaluation Metrics} % 1 - 1.5 P
\label{sec: sandbox environment}
\noindent
In this section we describe the objective of AL and also our sandbox environment.
The main setting we propose consists of two semi-synthetic OD datasets and down-scaled versions of standard object detectors leaving the detection mechanism unchanged.
Additionally, we introduce evaluations capturing different aspects of the observed AL curve.

\subsection{Active Learning}
\noindent
The term active learning refers to a setup (\cf \cref{fig: al cycle}) where %there is a limited amount of data $\mathcal{L}$ fully annotated towards a specific task 
only a limited amount of fully annotated data $\mathcal{L}$ is available together with a task-specific model.
% In addition, there is a pool or a stream of unlabeled data $\mathcal{U}$ from which the model selects ``actively''/queries those samples $\mathcal{Q}$ which are deemed informative/conducive towards test performance.
In addition, there is a pool or a stream of unlabeled data $\mathcal{U}$ from which the model queries those samples $\mathcal{Q}$ which are most informative.
Success of the query strategy is measured by observing that training on the new data pool leads to an increase of test performance higher than another method.
Afterwards, $\mathcal{Q}$ is annotated by an oracle (usually a human worker), added to $\mathcal{L}$ and the model is fine-tuned or fitted from scratch on $\mathcal{L}$ from where the cycle continues.
Before each acquisition step, the current model performance is evaluated which leads to graphs like the ones in \cref{fig: auc metric}.
The query step can take diverse algorithmic forms, see \cref{sec: related work} or \cite{settles.tr09}.

\subsection{Datasets}
\label{sec: nutshell datasets}
\noindent
% Common object detection tasks have a tendency to be selectively and loosely labeled, such as the COCO and Pascal VOC datasets.
We construct an OD problem by building a synthetic overlay to images from the real-world MS-COCO dataset (\cf \cref{fig: dataset generation}), which constitutes the data of our sandbox.
% \commentMS{\sout{The COCO background delivers a realistic, feature rich background against which foreground instances need to be discriminated from in this task.
% As foreground, we utilize two sets of categories: MNIST digits and EMNIST letters while deleting the COCO ground truth belonging to the background.}}
COCO images with deleted annotations provide a realistic, feature-rich background on which foreground objects shall be recognized.
As foreground, we utilize two sets of categories: MNIST digits and EMNIST letters.
% Both are randomly scaled, translated and sheared on a $320 \times 320$ pixel canvas.
We apply randomized coloration and opacity to foreground instances such that trivial edge detection becomes unfeasible.
Colors are drawn from HSV space uniformly by $(h, s, v) \sim U([0.0, 1.0] \times [0.05, 1.0] \times [0.1, 1.0])$ which yielded the best optical variability.
The chosen HSV-color is multiplied with the gray scale value of the original image avoiding the instances being monochromatic.
Opacity is similarly drawn uniformly $\alpha \sim U([0.5, 0.9])$ and in addition, we apply image translation, scaling and shearing to all numbers/letters.
The number of instances per background image is Poisson-distributed with mean $\lambda=3$.
% The number of instances per background canvas is Poisson distributed with mean $4$.
Tight bounding box (and instance segmentation) annotations can be obtained from the original transformed gray scale (E)MNIST images and the category label can be adopted.
% The semantic distributions in our datasets are similar to the distributions in the underlying image classification dataset.
Compared to simple detection datasets such as SVHN\cite{netzer2011reading}, the geometric variety in our datasets is more similar to those of OD benchmarks such as Pascal VOC or MS-COCO.
The reduction in the dataset complexity allows for the achievement of decent performance even for small architectures and leads to quickly converging training and low inference times.
% \commentMS{Especially the rapid convergence in the training process is crucial for saving time.}
% \HG{HG:Ich hätte gerne etwas mehr technische Details, wie zB die Anzahl der imputierten Ziffern/Buchstaben gewählt wurde und wie opacity und farbe randomisiert wurden. Haben wir über die Ziffer ein chargierende Färbung oder ist das monochom (und damit leicht zu erkennen)?}
In the following we use the terms ``MNIST-Det'' and ``EMNIST-Det'' to refer to our OD datasets.

\subsection{Models}
% \begin{table}
%     \centering
%     \caption{Exemplary object detection architectures employed during the experiments and associated number of weights.}
%     \begin{tabular}{l l c c} \toprule
%         Detector & Backbone & \# classes & \# params \\\midrule
%         RetinaNet & ResNet18 & 26 & 20.1M \\
%         Faster R-CNN & ResNet18 & 26 & 28.3M \\
%         YOLOv3 & Darknet20 & 26 & 10.3M \\
%         RetinaNet & ResNet50 & 20 & 36.5M \\
%         Faster R-CNN & ResNet101 & 20 & 60.2M \\
%         YOLOv3 & Darknet53 & 20 & 61.6M \\
%         \bottomrule
%     \end{tabular}
%     \label{tab: models and parameters}
% \end{table}

\begin{table}
    \centering
    \caption{Exemplary object detection architectures with backbone configurations employed in the experiments and associated number of parameters.}
    \resizebox{\columnwidth}{!}{
    \begin{tabular}{l | l c | l c} \toprule
        Detector & Backbone & \# params & Backbone & \# params \\\midrule
        RetinaNet & ResNet50 & 36.5M & ResNet18 & 20.1M \\
        Faster R-CNN & ResNet101 & 60.2M & ResNet18 & 28.3M \\
        YOLOv3 & Darknet53 & 61.6M & Darknet20 & 10.3M \\
        \bottomrule
    \end{tabular}
    }
    \label{tab: models and parameters}
\end{table}
\noindent
Modern OD architectures utilize several conceptually different mechanisms to solve the detection task.
Irrespective of the amount of accessible data, some applications of OD may require high inference speed while others may require a large degree of precision or some trade-off between the two. %\commentMS{or even both}.
The type of architecture as well, as the underlying detection mechanism are, however, disjoint to some degree from the depth of the feature extraction in the backbone.
The latter is mainly responsible for the quality and resolution of features.
We use this insight to down-scale architectures for AL by reducing the network depth while keeping the detection mechanism in the network's head unchanged.
Together with the simplified OD objectives from \cref{sec: nutshell datasets}, we obtain a well-performing, low-capacity and fast-inference setup to study AL with the same OD mechanisms that occur in practice.
\cref{tab: models and parameters} shows the choices for a YOLOv3\cite{redmon2018yolov3}, RetinaNet\cite{lin2017focal} and Faster R-CNN\cite{ren2015faster} setup, which we have made for our investigations together with the resulting number of parameters.
The number of parameters in the standard setup was reduced by up to a factor of around $6$ leading to a significant decrease in training and inference time.

% In order to facilitate testing modern object detectors as well as standard object detectors, we integrate our framework into the large MMDetection object detection toolbox.
% It has been continuously updated during the last couple of years and kept up with the latest developments of cutting edge object detection.
% Moreover, it allows for a high degree of modularity in terms of architectures, datasets and training schedules and also offers pre-trained backbones and models which can and, usually are, used for transfer training or finetuning of object detectors.
% This not only drastically accelerates the training phase but also constitutes a realistic setup for any active learning endeavors.
% We implement the baselines used in this work for three architectures which have been widely employed in the active learning literature.
% In the forward pass of a neural network, inference time is strongly influenced by the depth of the architecture as well, as by the type of layers utilized.
% We, therefore, propose to reduce the depth of object detectors to be better suited to the reduced-complexity object detection tasks introduced in the last paragraph.
% This leads to a far smaller number of parameters to be fitted during training and to a lower required number of optimizer iterations.

\subsection{Active Learning Methods in Object Detection}
\label{sec: al and methods in od}
\noindent
For the construction of baseline AL methods, we focus on querying whole images instead of single bounding boxes. 
The latter would introduce an additional dimension of complexity where unlabeled image regions need to be ignored.
The frequently used uncertainty-based query strategies from image classification, such as entropy, probability margin, MC dropout, and mutual information, determine prediction-specific but not image-wise uncertainties.
Hence, we introduce an aggregation step to obtain image-wise query scores.

For a given image $\mathbf{x}$, a neural network predicts a fixed number $N$ of bounding boxes $\hat{b}_{\mathbf{x}}^{(i)}=\{x_{\min}, y_{\min},x_{\max},y_{\max},s,p_1,\ldots,p_C\},\ i=1,\ldots,N$, where $x_{\min}, y_{\min},x_{\max},y_{\max}$ represent the localization, $s$ the objectness score (or analog) and $p_1,\ldots,p_C$ the class probabilities for the $C$ classes of the prediction.
Only the set of boxes that remain after non maximum suppression (NMS) and score thresholding are used to determine prediction uncertainties.
Already the choice of the parameters for the NMS and the score threshold influence the queries, since they decide which (uncertain) predictions remain.

Given a prediction $\hat{b}$  we compute its classification entropy $H(\hat{b}) = - \sum_{c = 1}^C p_c \cdot \log(p_c)$ and its probability margin score $PM(\hat{b}) = (1 - [p_{c_{\max}} - \max_{c \neq c_{\max}} p_c])^2$.
Here, $c_{\max}$ denotes the class with the highest predicted probability.
When dropout is implemented in the architecture, Monte-Carlo Dropout samples can be drawn at inference time where the output of the same anchor box $\hat{b}_1,\ldots,\hat{b}_K$ are sampled $K$ times . 
The final prediction under dropout is then the arithmetic mean $\overline{\hat{b}} = \tfrac{1}{K} \sum_{i = 1}^K \hat{b}_i$.
The Monte-Carlo mutual information is estimated by $\mathit{MI}(\hat{b}) =H(\overline{\hat{b}}) - \overline{H(\hat{b})}$ with the second term being the average entropy of the individual samples.
We also regard the maximum feature standard deviations within $\hat{b}$ by first standardizing variances over all query predictions to treat localization and classification features on the same footing.
The dropout uncertainty is then $D = \max_{\phi \in \{x_{\min}, y_{\min},x_{\max},y_{\max}, s, p_1, \ldots, p_C\}} \sigma(\widetilde{\phi})$.
Note that for all these methods, uncertainty is only considered in the predicted foreground instances.
Since the uncertainty-based selection strategies only determine prediction-based uncertainties, either the sum, average, or maximum is taken over predicted instances to obtain a final query score for the image.
Summation, for instance, tends to prefer images with a large amount of instances while averaging is strongly biased by the thresholds (imagine considering many false positive predictions which could be filtered by a higher threshold).
In the presented experiments, we use summation and further aggregations are investigated in the appendix.
Additionally, we also regard random acquisition as a completely uninformed query baseline.

% \MR{\textbf{MR: gibt es zu den Aggregationen eine Studie im Appendix? Wir sollten schon anmerken, dass wir es untersucht haben}}
% Similarly, 
Diversity-based methods make use of latent activation features in neural networks which heavily depend on the OD architecture.
Since purely diversity-based methods have been less prominent in the literature, we focus on the more broadly established uncertainty baselines. % in our investigations. \commentMS{wollen wir das dann erwähnen?}

\subsection{Evaluation}
\label{sec: evaluation}
\noindent
% \textbf{Model performance.}
\paragraph{Model Performance.}
In the literature, methods are frequently evaluated by counting the number of data samples needed to cross some fixed reference performance mark.
For OD, performance is usually measured in terms of the $\map_{50}$ metric for which there is a maximum value known when training on all available data.
In AL then, some percentage of this maximal performance, \eg, $90\%$ needs to be reached with as few data points as possible.
Collecting performance over amount of queried data gives rise to curves such as the one in the top right of \cref{fig: al cycle} which we call \emph{AL curves} in the following.

\noindent
% \textbf{Counting annotations.}
\paragraph{Counting Annotations.}
``Amount of training data'' usually translates to the number of queried images which is set as a hyperparameter and fixed for each method and AL cycle.
Considering that in the annotation process, each bounding box needs to be labeled and there tends to be high variance in the number of instances per image in common benchmark datasets, it is not clear whether to measure annotated data in terms of images or instances.
Therefore, we stress that the scaling of the $t$-axis is important especially in instance-wise prediction tasks such as OD.
% \finding{However, the \emph{mechanical effort} in the annotation process is instead always proportional to the number of instances queried.}
% Therefore, we also strongly recommend to include the analog curves of performance over number of queried instances (instead of queried images) for OD to get a more transparent view of each method.
% Averaging of individual seeds of one and the same experiment need to be done with care as they cannot immediately be averaged at each ($t$-)evaluation point.
% However, linear interpolations of the curves up to the maximum number of ever queried instances can be averaged, so we obtain point-wise standard deviations of the performance.
% Typically this is information present anyway during the experiment and comes at no additional cost.
Both views, counting images or instances, can be argued for.
Therefore, we evaluate the performance of each result not only based on the acquired images, but also transform the $t$-axis under the curves to the number of annotated instances.
By linear interpolation between query points and averaging of individual seeds of the same experiment, we obtain image- or instance-wise standard deviations of the performance.

\noindent
% \textbf{Area under AL curves.}
\paragraph{Area under AL Curves.}
\begin{figure}
    \centering
    \begin{overpic}[width=0.8\linewidth]{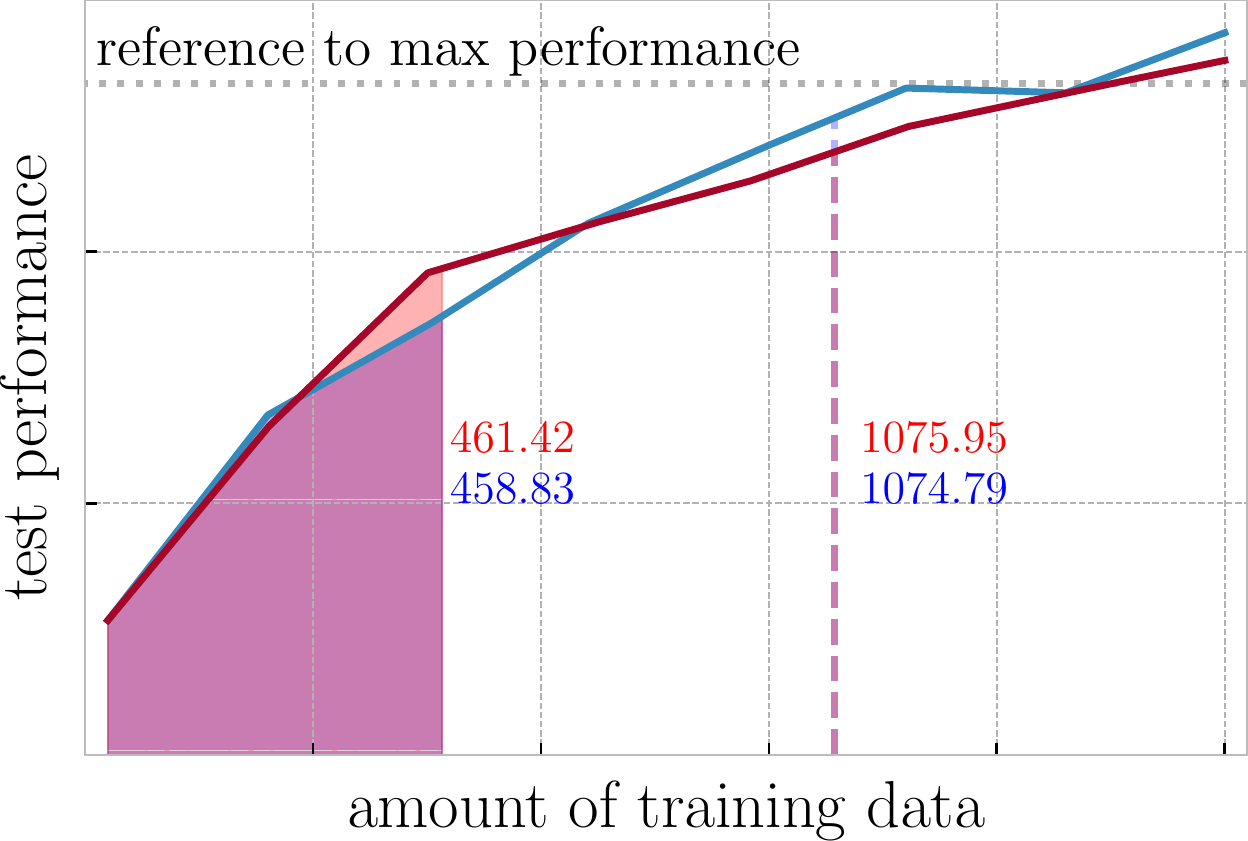}
    \put (36,8){$t_1$}
    \put (68,8){$t_2$}
    \end{overpic}
    \caption{Area under AL curve ($\auc$) metric at different stages of an AL curve for two different query strategies (averaged, taken from \cref{fig: active learning curves voc bdd}).
    }
    \label{fig: auc metric}
\end{figure}
In light of the complexity of the AL problem, we propose the area under the AL curve ($\auc$). 
It constitutes a more robust performance metric compared to (horizontal or vertical) cross sections through the learning curves.
\cref{fig: auc metric} shows two AL curves and corresponding $\auc$ metrics at two distinct points $t_1$ and $t_2$.
Note that in practice, there is no fixed maximal dataset size for which a maximum performance exists.
Then, the AL experiment may end and be evaluated at any given vertical section of $t$ points of training data.
Knowing the maximal performance (or the $90\%$ mark shown in \cref{fig: auc metric}) may lead to wrong conclusions in the presented case which is taken from the scenario in \cref{fig: active learning curves voc bdd}.
% \commentMS{\sout{Ending the experiment at $t_1$, the left of the two chosen evaluation points in \cref{fig: auc metric} clearly determines the red curve (which also has a higher $\auc$) as preferable.
% The right evaluation point $t_2$ seems to tell a different story with the blue line in front in terms of pure detection performance.
% However, the prior history is not congruent with, say, the numbers reflected in the pure performance evaluation at that point.
% Observing the $\auc$ instead of the raw performance at $t_2$ indicates, that the red curve is still slightly ahead of the blue curve although the difference is marginal here.
% This is in line with our qualitative feeling of the curves when regarded up to $t_2$.
% With the trend of blue showing higher test performance than red continuing, the $\auc$ will reflect that behavior and eventually favor the blue curve.}}
Ending the experiment at $t_1$ clearly determines the red curve (which also has a higher $\auc$) as preferable.
Ending the experiment at $t_2$ favors the blue curve by just looking at the actual test performance. 
However, the $\auc$ still favors the red curve, since it takes the complete AL curve into account.
This is in line with our qualitative feeling of the curves when regarded up to $t_2$.
Note that we use the $\auc$ metric for calculating rank correlations in \cref{sec: experimental representativity}.
The raw results of the $\auc$ metric are shown in the appendix.

\section{Experiments}
\label{sec: developing al}
\noindent
In this section, we present results of experiments with our sandbox environment as well as established datasets, namely Pascal VOC and BDD100k, in the following abbreviated as VOC and BDD.
We do so by presenting AL curves, summarizing benchmark results and discussing our observations for different evaluation metrics in \cref{sec: benchmark results}. 
We then show in \cref{sec: experimental representativity} quantitatively that our sandbox results generalize to the same extent to VOC and BDD as results obtained on those datasets generalize between each other. 
In other words, we demonstrate the dataset-wise representativity of the results obtained by our sandbox. 
Thereafter, this is complemented with a study on the computational speedup in \cref{sec: compute time} when using the sandbox instead of VOC or BDD.

\noindent
\paragraph{Implementation.}
\begin{table}[]
    \centering
    \caption{Maximum $\map_{50}$ values achieved by the models in \cref{tab: models and parameters} on the respective datasets (standard-size detectors on VOC and BDD; sandbox-size on (E)MNIST-Det).
    The entire available training data is used.
    }
    \resizebox{0.85\linewidth}{!}{
    \begin{tabular}{l|c c c}
    \toprule
     & YOLOv3 & RetinaNet & Faster R-CNN \\
     \midrule
    MNIST-Det & 0.962 & 0.908 & 0.937 \\
    EMNIST-Det & 0.959 & 0.919 & 0.928 \\
    Pascal VOC & 0.794 & 0.748 & 0.797 \\
    BDD100k & 0.426 & 0.464 & 0.525\\
    \bottomrule
    \end{tabular}
    }
    \label{tab: max performance}
\end{table}
\begin{table*}[]
    \centering
    \caption{Amount of queried images and bounding boxes necessary to cross the $90\%$ performance mark during AL.
    Lower values are better.
    Bold numbers indicate the lowest amount of data per experiment and underlined numbers are the second lowest.
    }
    \resizebox{\textwidth}{!}{
    \input{tables/num_images_queried}
    }
    \label{tab: query amounts}
\end{table*}
We implemented our pipeline in the open source MMDetection\cite{mmdetection} toolbox.
In our experiments for VOC, $\mathcal{U}$ initially consists of ``2007 train'' + ``2012 trainval'' and we evaluate performance on the ``2007 test''-split.
At initialization, we randomly sample $\mathcal{L}$ as a small portion of $\mathcal{U}$.
When tracking validation performance to assure convergence, we evaluate on ``2007 val''.
Since BDD is a hard detection problem, we filtered frames with ``clear'' weather condition at ``daytime'' from the ``train'' split as initial pool $\mathcal{U}$ yielding $12,\!454$ images.
We apply the same filter to the ``val'' split and divide it in half to get a test dataset ($882$ images for performance measurement) and a validation dataset ($882$ images for convergence tracking).
For the (E)MNIST-Det datasets we generated $20,\!000$ train images, $500$ validation images and $2,\!000$ test images.
For reference, we collect in \cref{tab: max performance} the achieved performance of the respective models for each dataset which determines the $90\%$ mark investigated in our experiments.
Details on the architectures and concrete parameters for the AL experiments conducted are provided in the appendix.

\subsection{Benchmark Results}
\label{sec: benchmark results}

\begin{figure}
    \begin{center}
        \includegraphics[width=\linewidth]{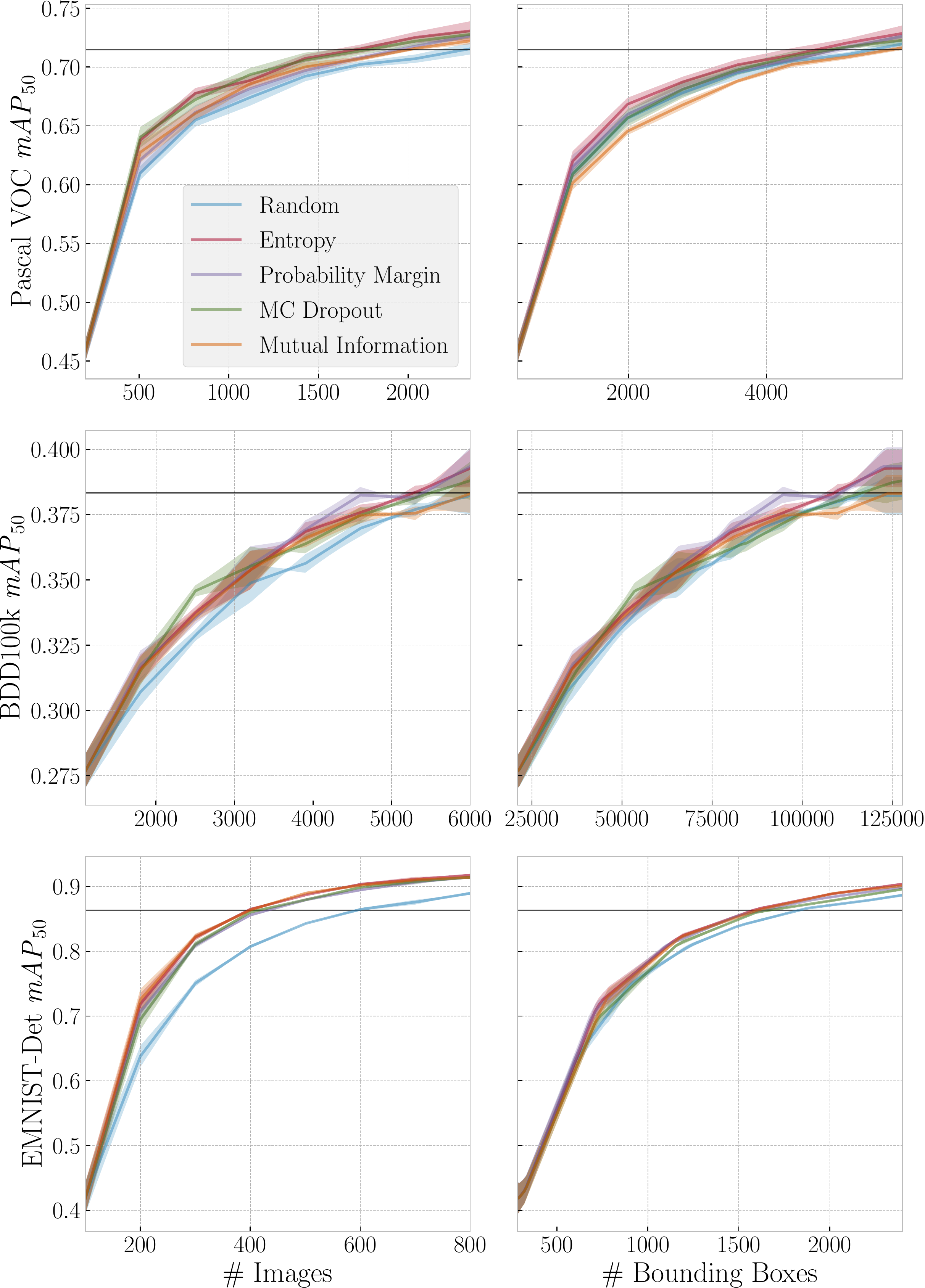}
    \end{center}
    \caption{Comparison of YOLOv3 AL curves on three different datasets.}
    \label{fig: active learning curves voc bdd}
\end{figure}
\begin{figure*}
    \centering
    \begin{subfigure}[c]{0.5\textwidth}
        \centering
        \subcaption{YOLOv3}
        % \includegraphics[height=\plotheight]{images/correlations_yolo_images.pdf}
        % \hspace{-0.6cm}
        % \includegraphics[height=\plotheight]{images/correlations_yolo_boxes.pdf}
        \rotatebox{90}{\qquad\quad$\map_{50}$}
        \includegraphics[height=3.05cm]{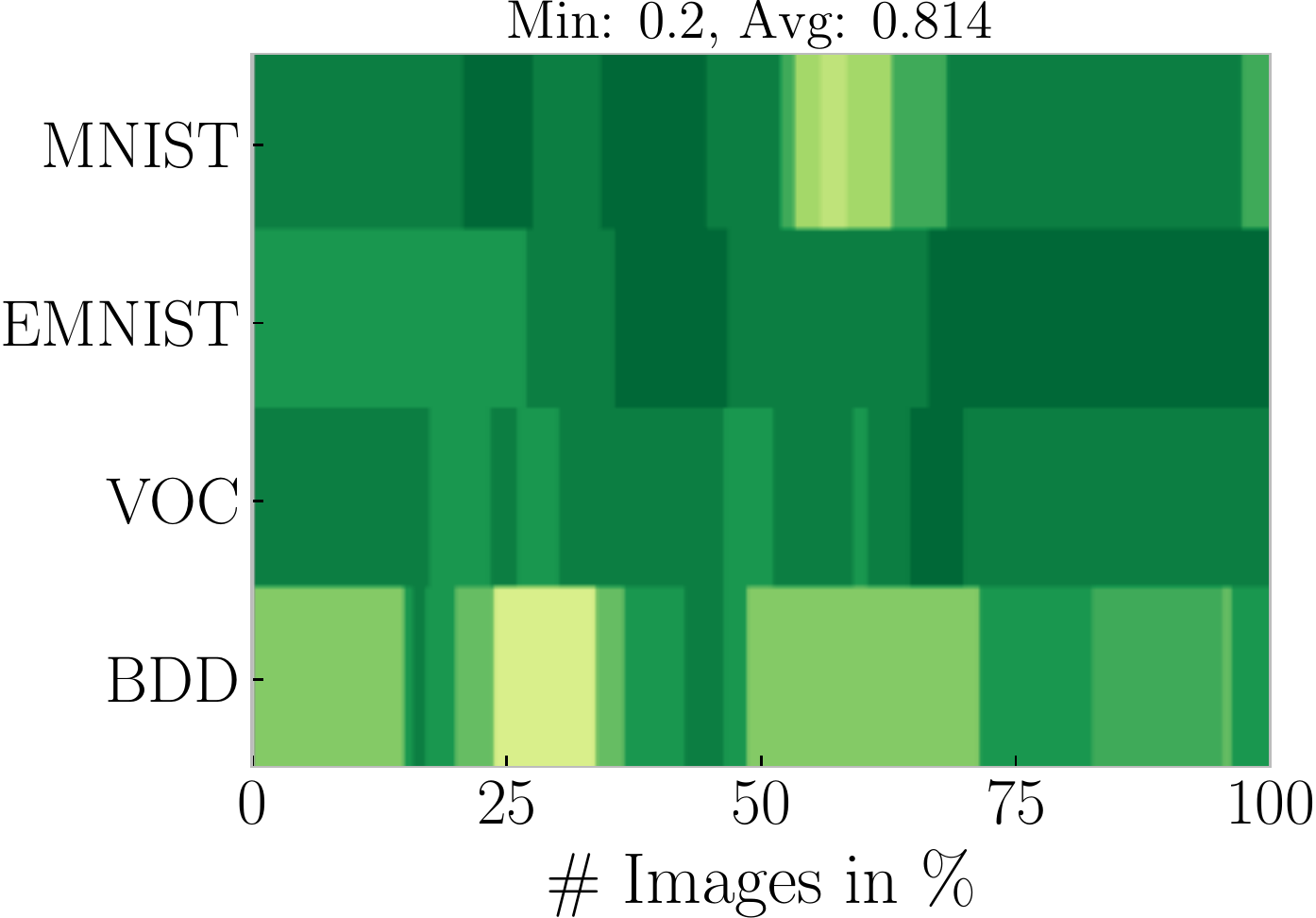}
        \includegraphics[height=3.05cm]{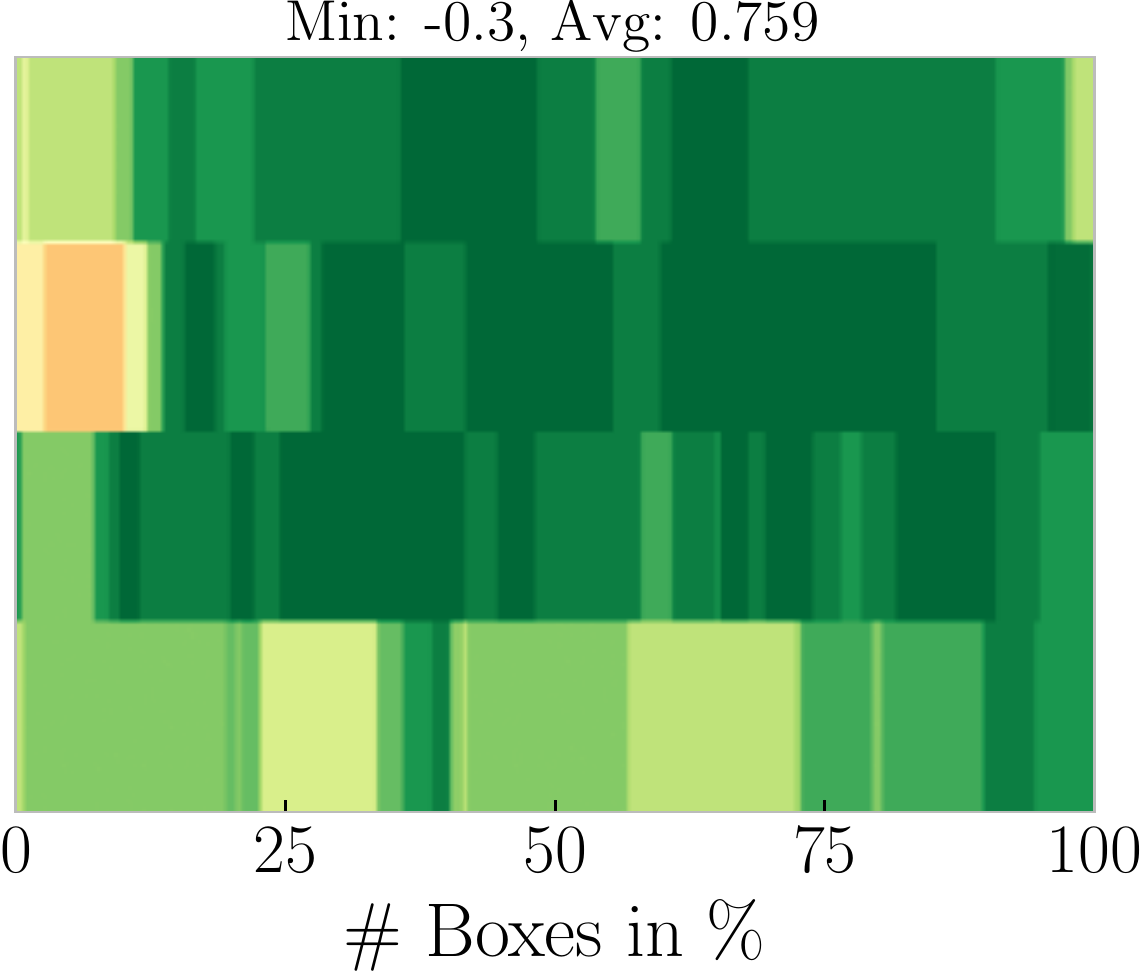}
        \newline
        \rotatebox{90}{\qquad\quad$\auc$}
        \includegraphics[height=3.05cm]{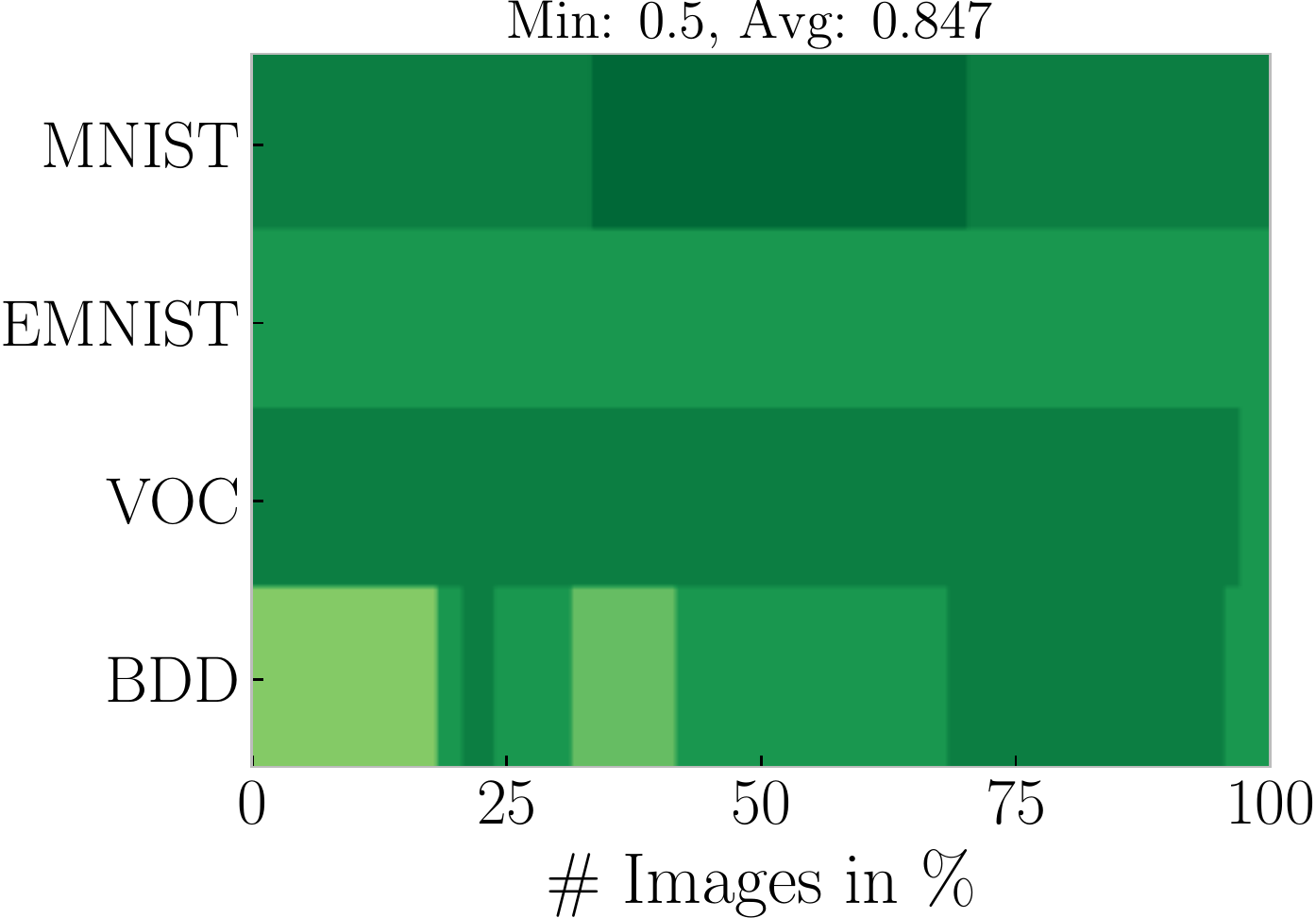}
        \includegraphics[height=3.05cm]{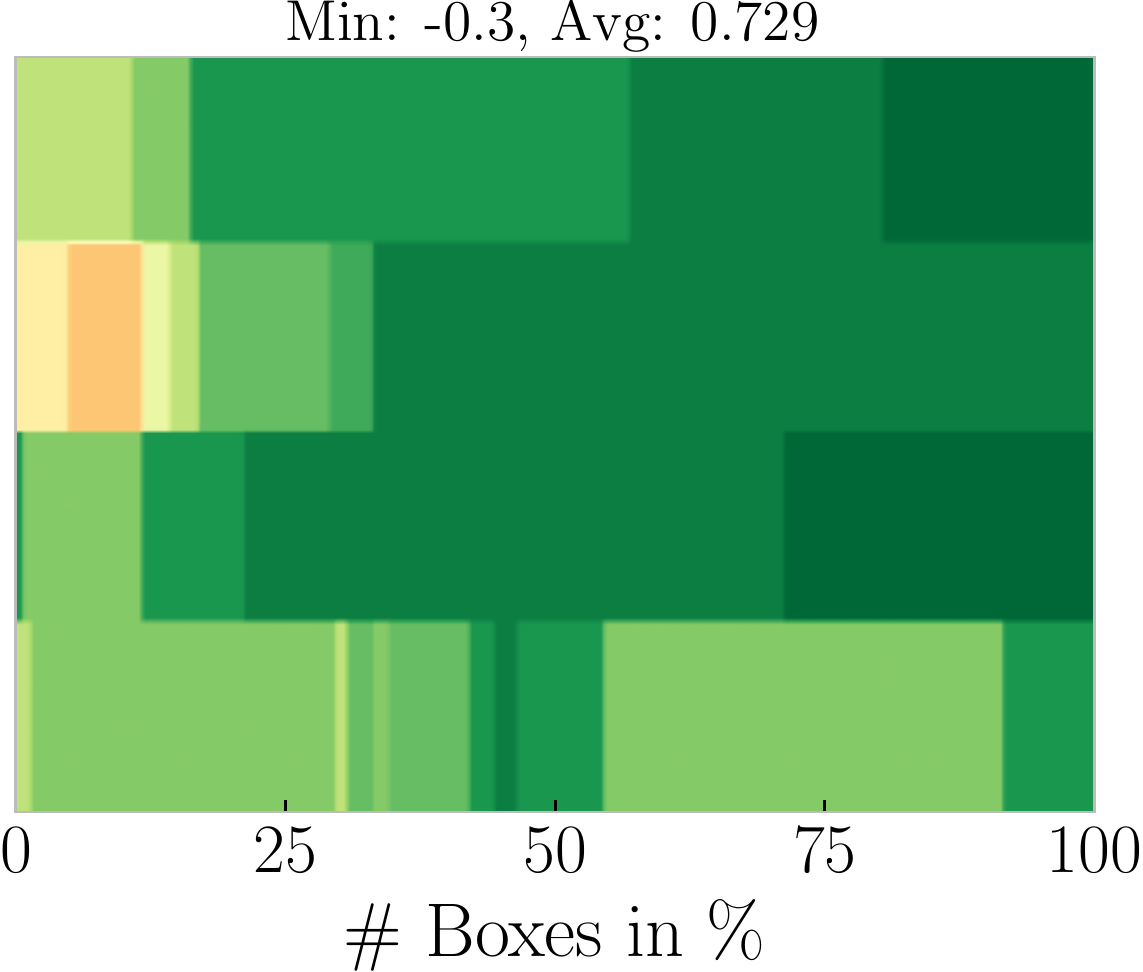}
    \end{subfigure}
    % \hspace{-0.6cm}
    \begin{subfigure}[c]{0.45\textwidth}
        \centering
        \subcaption{Faster R-CNN}
        % \includegraphics[height=\plotheight]{images/correlations_frcnn_images.pdf}
        % \hspace{-0.6cm}
        % \includegraphics[height=\plotheight]{images/correlations_frcnn_boxes_bar.pdf}
        \includegraphics[height=3.05cm]{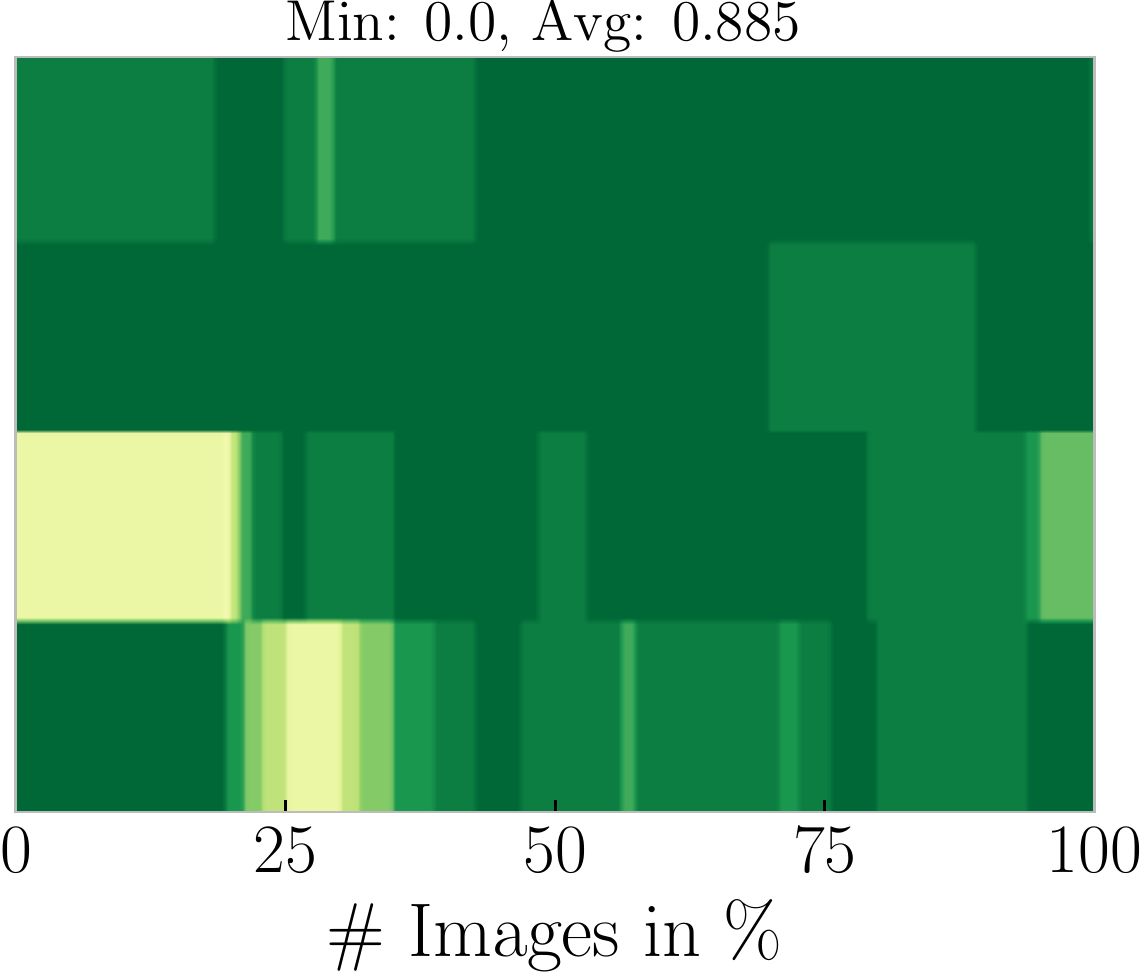}
        \includegraphics[height=3.05cm]{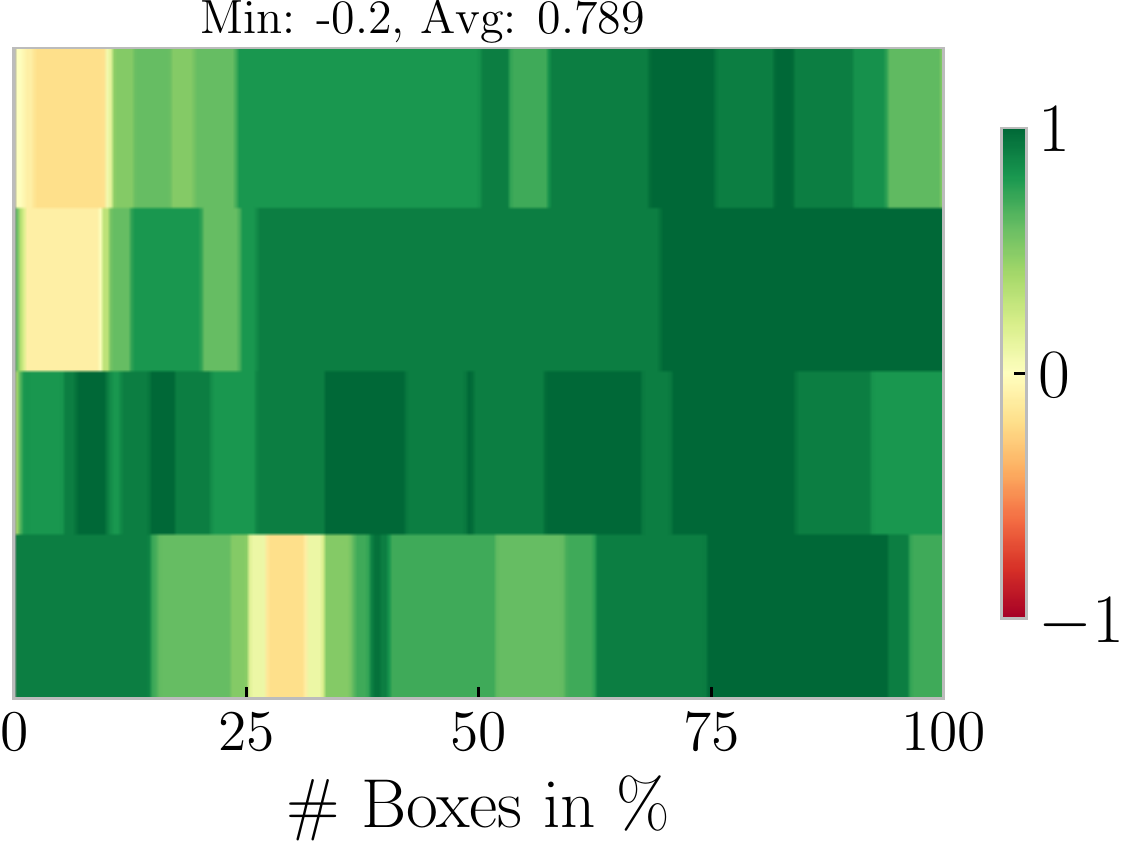}
        \includegraphics[height=3.05cm]{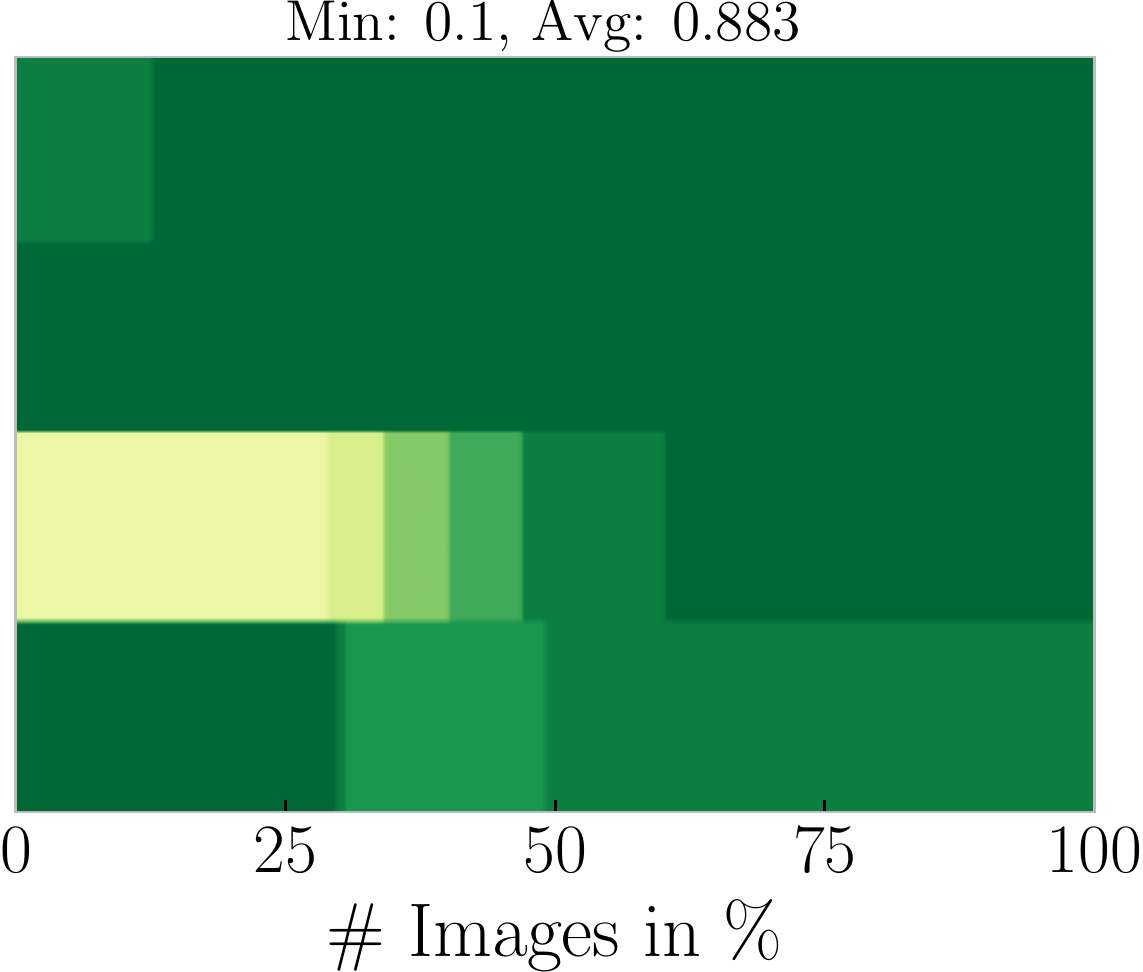}
        \includegraphics[height=3.05cm]{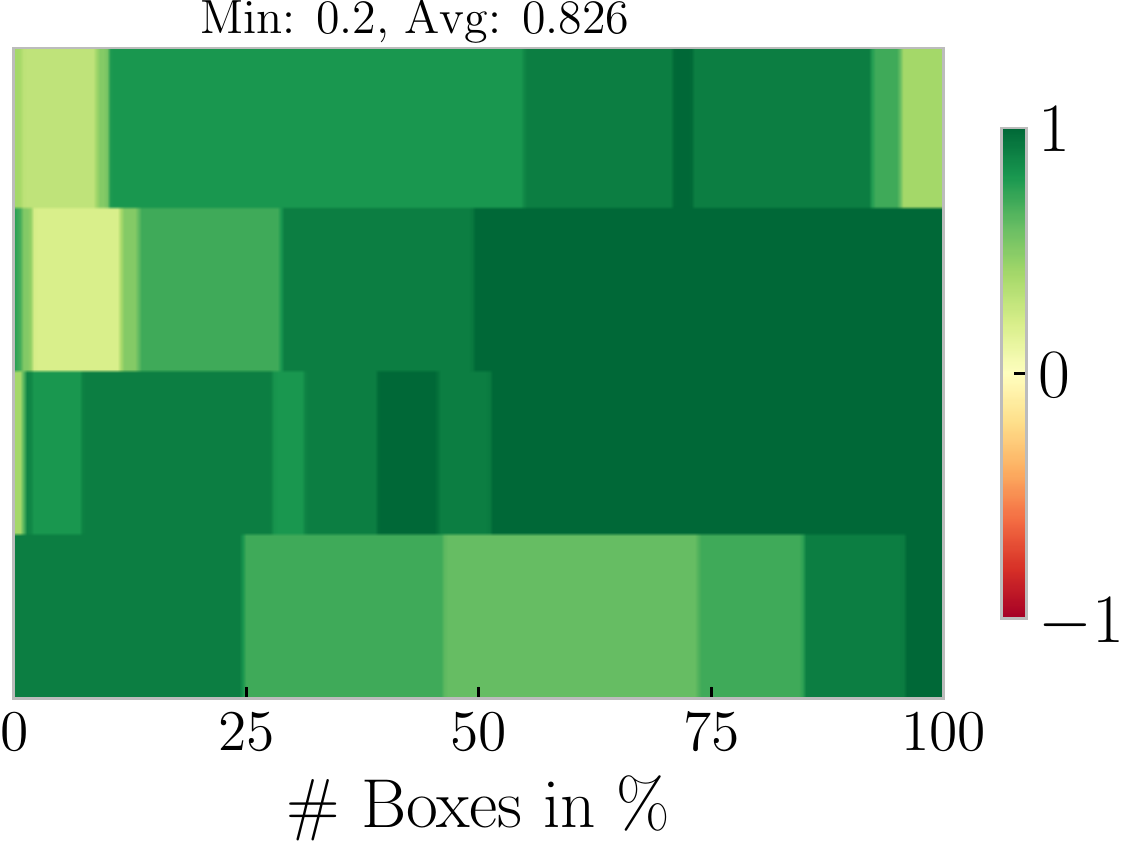}
    \end{subfigure}
    \caption{Curves of rank correlations between the cumulative $\auc$ and the final rankings at the $90\%$ max performance mark.}
    \label{fig: correlation histories}
\end{figure*}
\noindent
% Seemingly similar AL setups can exhibit vastly different behavior in evaluation.\HG{Den letzten Satz weglassen?}
We first investigate differences in AL results w.r.t.\ the datasets where we fix the architecture.
The comparison uses the YOLOv3 detector on two standard OD benchmarks (VOC and BDD) and our EMNIST-Det dataset.
Our comparison includes the five query baselines described in \cref{sec: al and methods in od}. 
We obtain AL curves averaged over four random initializations and evaluated in terms of queried images as well as in terms of queried object instances, respectively. 
\cref{fig: active learning curves voc bdd} shows the test performance curves with shaded regions indicating point-wise standard deviations.
The left panels show performance according to queried images while the right panels show the same curves but according to queried instances.
We observe that the uncertainty-based query strategies tend to consistently outperform the Random query in image-wise evaluation.
However, when regarding the number of queried bounding boxes, the situation is far less clear.
For EMNIST-Det, the difference between the Random and the uncertainty-based queries decreases substantially, such that only a marginal difference is visible.
In VOC and BDD, the Random baseline falls roughly somewhere in-between the uncertainty baselines in instance-wise evaluation.
This indicates that greedy acquisition of images with highest sum of uncertainty tends to query images with a large amount of ground truth boxes.
Obtaining many ground truth signals improves detection performance in these cases, while the query of large amounts of boxes gives rise to a higher annotation cost on the right panels.
From this observation, we conclude that comparing AL curves based only on the number of acquired training images gives an incomplete impression of the method and the associated annotation costs.
In addition to the image-wise evaluation, an instance-wise evaluation should be considered.
Note also, that the AL curves for EMNIST-Det and VOC have a smoother progression than those for BDD.
We attribute this finding to the fact that BDD is a far more complicated detection problem which includes a large amount of small objects. 
However, the fluctuations in the AL curves on BDD tend to average out as we consider the $\auc$ of the AL curve as evaluation metric. 
This becomes clear in light of the results in \cref{sec: experimental representativity}, where we study the generalization across datasets in terms of the $\auc$ of the AL curve.

%\MR{However, the question which method dominates which other method over the AL curve and how this generalizes across datasets might be of greater interest. To this end, we consider rankings} of the methods for a fixed data set and draw comparisons to the rankings of the other datasets in \cref{sec: experimental representativity}.}
%\HG{HG: wie wird das festgestellt -- verweise hier  rank-correlation auswertungen (oder beides) in der Figure 5.}
%\TR{This will be further investigated in \cref{sec: experimental representativity}.}
%\commentMS{\sout{The difference in scaling of both axes indicates a far more complicated detection problem in BDD100k which includes a large amount of small objects.
%We conclude that the viability of AL methods can be highly context- and dataset-dependent.}}
%\HG{HG: Wie setht diese Aussage zu der sandbox-Argumentation? Ist das nicht widersprüchlich?}

In \cref{tab: query amounts} we show extended benchmark results.
For each detector to reach $90\%$ of the maximum performance, the table shows the number of queried images required, resp.\ the number of bounding boxes for each method.
These numbers were acquired as the average of four runs.
We see the rankings per experiment which often favor the Entropy baseline, however, the overall rankings are rather unstable throughout the table.
Note in particular, that for the arguably hardest detection problem BDD, the Random baseline beats the Mutual Information baseline for the YOLOv3 detector.
In the analog setting for Faster R-CNN the image-wise margin of the Mutual Information becomes slim.
This observation also holds for instance-wise evaluation and is even more pronounced where in six cases, Random beats an informed query baseline.
Results for the RetinaNet detector can be found in the appendix.
In light of the discussion in \cref{sec: evaluation}, we conclude that in order to assess the viability of a query strategy, AL curves should be viewed from both angles: performance over number of images and over number of bounding boxes queried.

\subsection{Generalization of Sandbox Results}
\label{sec: experimental representativity}
\noindent
Instead of evaluating the pure performance at each AL step we have proposed in \cref{sec: evaluation} to compute the corresponding $\auc$ as a more robust metric of AL performance.
With respect to the final method ranking at some fixed reference mark, we compute Spearman rank correlations with the $\map_{\!50}$ metric at each point $t$. % which is frequently used to assess object detection performance.
We compare these with the analogous correlations with the respective $\auc$ at each point.
\cref{fig: correlation histories} shows intensity diagrams representing the rank correlations both, in terms of image-wise and instance-wise evaluation.
The reference mark we chose is $90\%$ of the maximum $\map_{\!50}$ obtained (\cf \cref{tab: max performance}).
The $t$-axes are normalized to the maximum number of images, resp.\ bounding boxes queried and the color indicates the Spearman correlation of the rankings.
Red represents negative correlation (\ie partial to full inversion of the observed ranking) while green means positive correlation (\ie a higher degree of similarity of the rankings).
In \cref{fig: correlation histories} both, $\map_{50}$ and $\auc$ show overall high correlation with the method ranking, especially towards the end of the curves.
We see that the correlations for $\auc$ fluctuates far less.
Moreover, the average correlation across entire AL curves tends to be larger for the $\auc$ metric than $\map_{\!50}$.
% Since in the real-world application of active learning, the reference mark of $x\%$ of the maximal performance is not accessible, we investigate an alternative performance metric of acquisition strategies.
% One point of motivation hereby is the difference between early versus late performance gains which may be context- or application-dependent, \ie, following the question ``How hard is it really, to obtain a single annotation?''.
% We investigate the cumulative $\auc$ scores (integrated up to $n$ images or bounding boxes) of different active learning histories and investigate the rank correlations at each $n$ with the final ranking of reaching the mark of $90\%$ max performance in each respective scaling of the abscissa.
% \cref{fig: correlation histories} shows such rank correlations for different datasets and two types of object detection architectures.
% The color scaling reaches from red (inverse ranking compared with $90\%$ max performance ranking) to green (same ranking).
% We see overall high correlation in both, image-wise and instance-wise evaluation with a tendency towards especially high correlations at the end of the history.
Note that the final ranking of either method does not need to be perfectly correlated with the $90\%$ max performance ranking for two reasons.
Firstly, the latter does not take into consideration early performance gains and secondly, the $90\%$ max performance ranking is a horizontal section through the curves while $\map_{\!50}$ and $\auc$ are vertical sections.
For a more quantitative evaluation, we show minimal and average correlation over all curves in \cref{fig: correlation histories}.
We observe overall high averages (upwards of $0.72$) throughout the curves.
The minimum $\auc$ correlation tends to be higher than the minimum $\map_{50}$ correlation.
We conclude that $\auc$ tends to be highly correlated with the $90\%$ max performance ranking and is more stable w.r.t.\ $t$ than $\map_{50}$.

\begin{figure*}
    \centering
    \def\plotheight{4cm}
    \begin{subfigure}[c]{0.49\textwidth}
        \centering
        \subcaption{YOLOv3}
        \includegraphics[height=\plotheight]{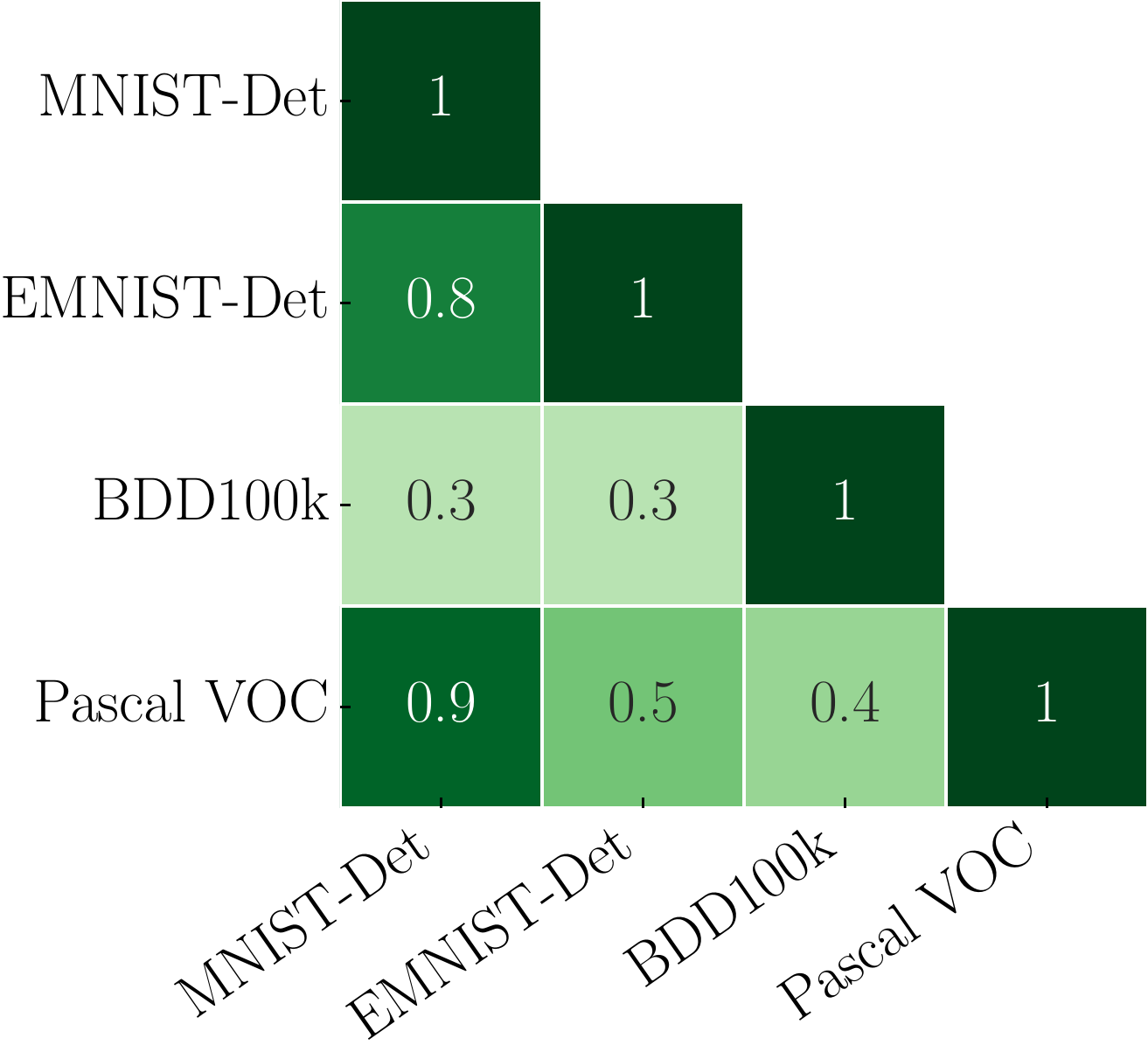}
        \hspace{-0.6cm}
        \includegraphics[height=\plotheight]{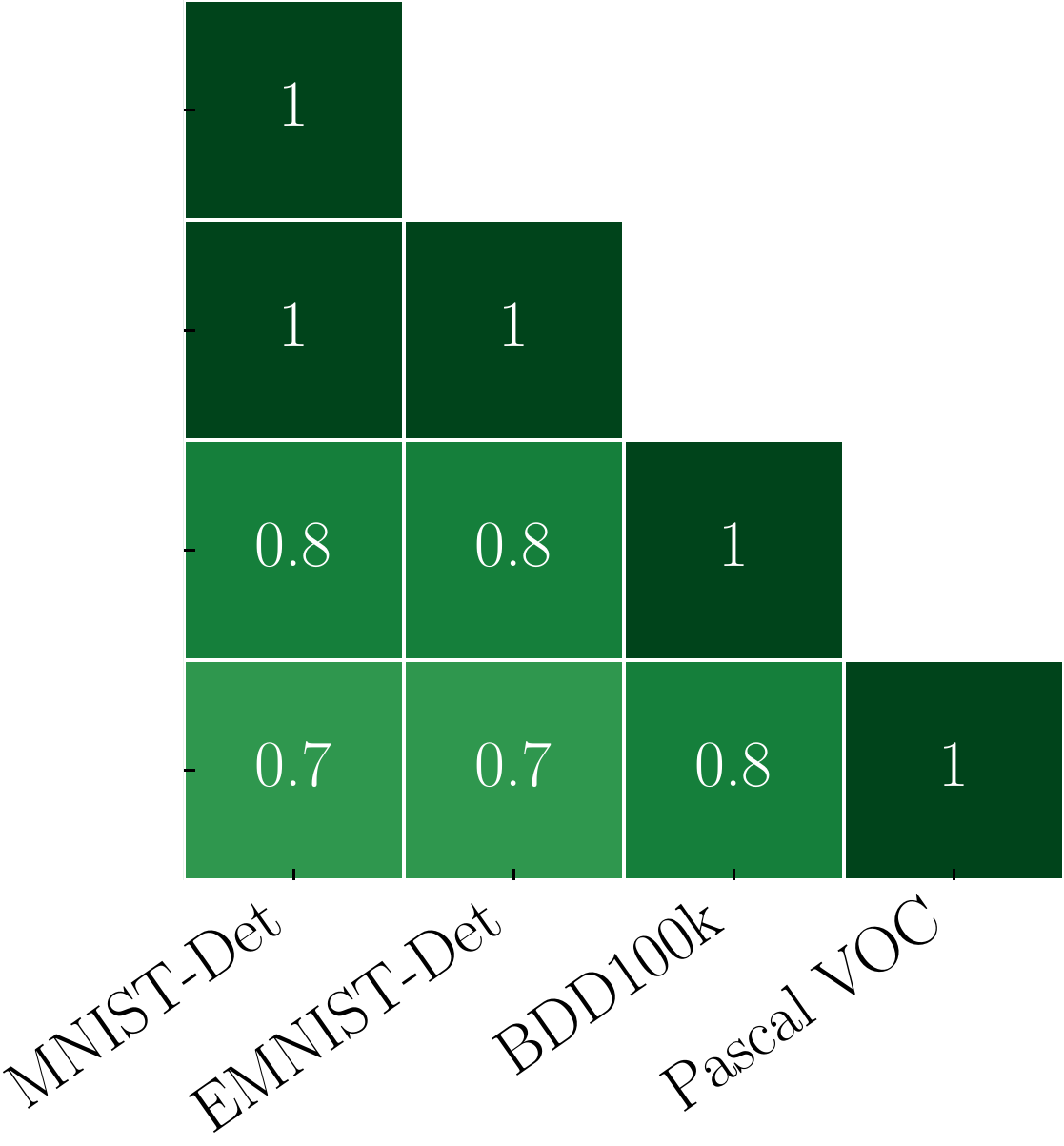}
    \end{subfigure}
    \hspace{-0.6cm}
    \begin{subfigure}[c]{0.49\textwidth}
        \centering
        \subcaption{Faster R-CNN}
        \includegraphics[height=\plotheight]{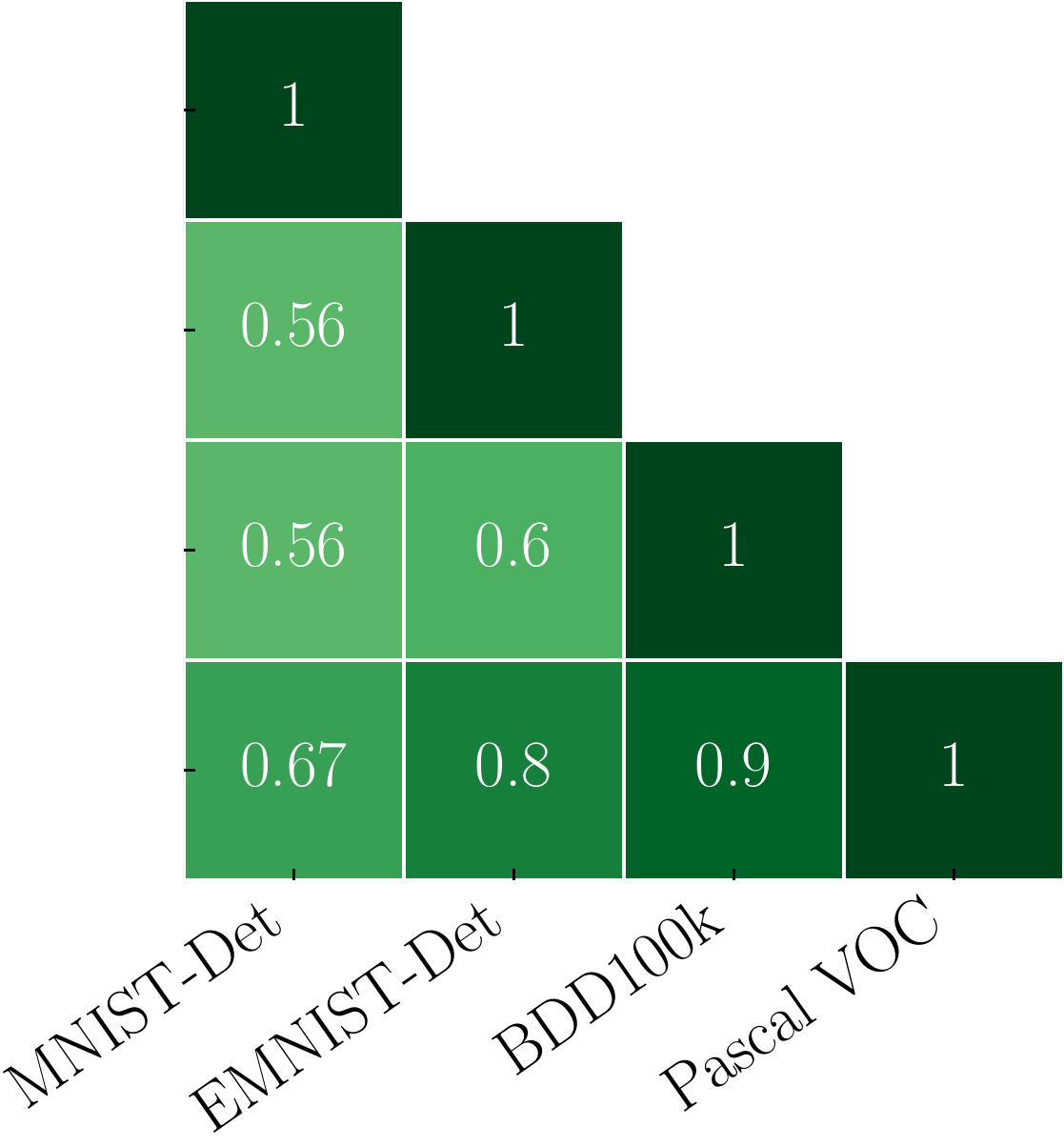}
        \hspace{-0.6cm}
        \includegraphics[height=\plotheight]{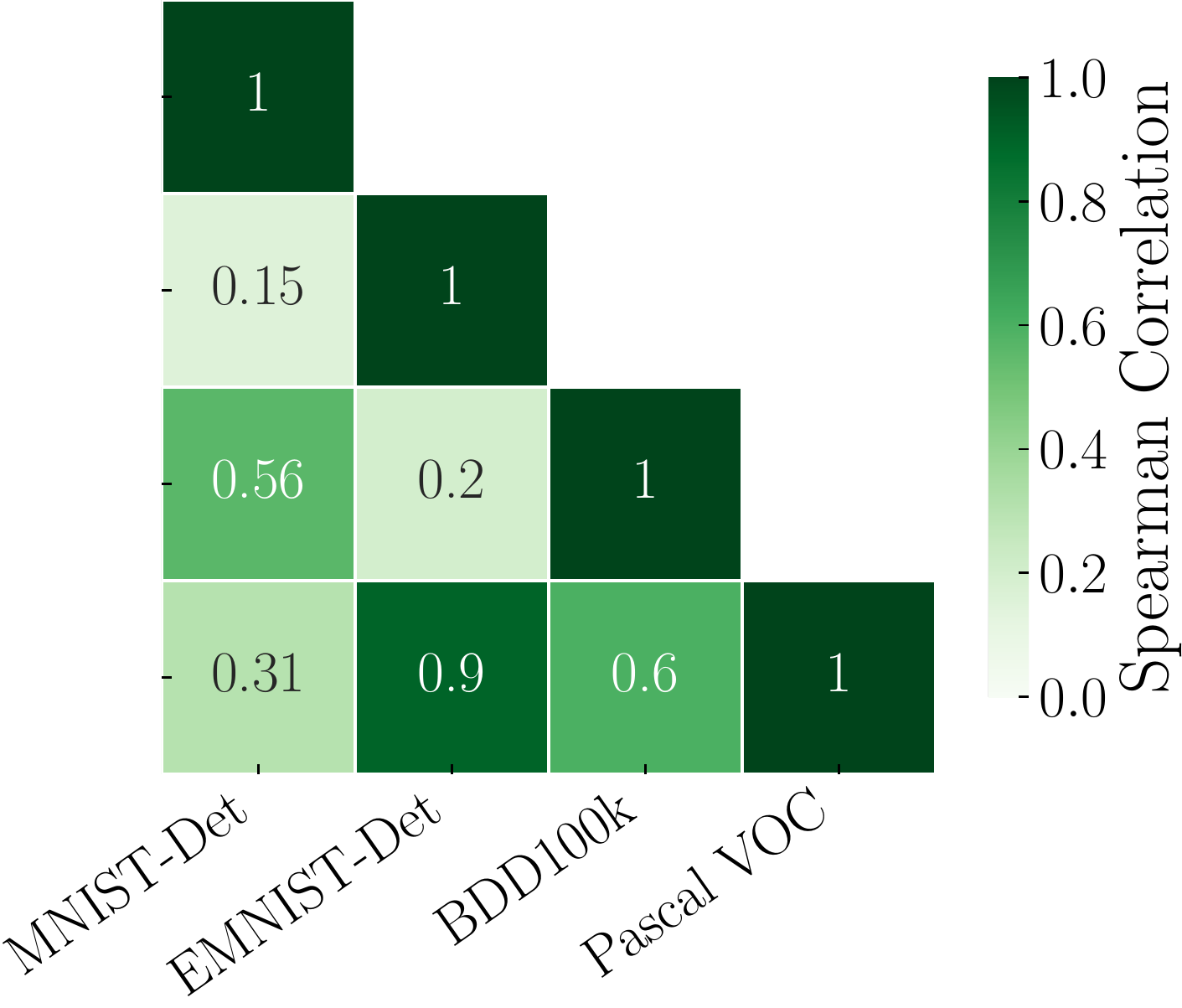}
    \end{subfigure}
    \caption{Ranking correlations between $\auc$ values for different detectors, left: image-wise; right: instance-wise evaluation.}
    \label{fig: rank correlations between datasets}
\end{figure*}
% Our proposition of developing and investigating simplified AL problems to minimize interference with the complexity of benchmark datasets requires an investigation of the comparability between our sandbox setting and standard AL settings in OD.
Next, we study how comparable AL experiments are between the sandbox setting and full-complexity problems (VOC and BDD).
% We investigate whether active learning in object detection in a simplified setting (like EMNIST-Det) is comparable to investigating a full-complexity problem (\eg, Pascal VOC or BDD100k with a very deep architecture).
To this end, we consider the cross-dataset correlations of the $\auc$ score when fixing the detection architecture.
\cref{fig: rank correlations between datasets} shows correlation matrices for image- and instance-wise evaluation.
% \sout{We show correlation matrices for image- and instance-wise evaluation in \cref{fig: rank correlations between datasets}.}
When comparing to the VOC-BDD correlations, the MNIST-Det and EMNIST-Det method rankings tend to be similarly correlated with either one.
Consider, for instance, the image-wise correlations on VOC.
For the YOLOv3 detector, both MNIST-Det and EMNIST-Det have higher correlation with VOC than BDD does.
Meanwhile, the BDD-MNIST-Det (resp.\ EMNIST-Det) correlation is a bit lower than VOC-BDD but similar.
For Faster R-CNN, the correlation VOC-BDD is large with $0.9$, however, both EMNIST-Det-VOC and EMNIST-Det-BDD have correlations of $0.6$ and $0.8$ which are reasonably high.
For all correlations of VOC with BDD, replacing either one of the datasets with MNIST-Det or EMNIST-Det results in similar correlation.
From this, we conclude that comparing methods in the simplified setting yields a similar amount of information about the relative performances of AL as the full-complexity setting.
% Again, the evaluation for RetinaNet can be found in the appendix. \MR{\textbf{MR: auch wenn die Abkuerzungen recht klar sind, wollt ihr sie vielleicht zu Beginn der Experimente einfuehren und dann konsistent verwenden?}}

\subsection{Compute Time}
\label{sec: compute time}
\begin{figure}
    \centering
    \resizebox{0.8\linewidth}{!}{
    \input{figures/timing_plot}
    }
    \caption{Utilized time for one AL step (training until convergence + evaluating the query) for investigated settings in hours.}
    \label{fig: runtime}
\end{figure}
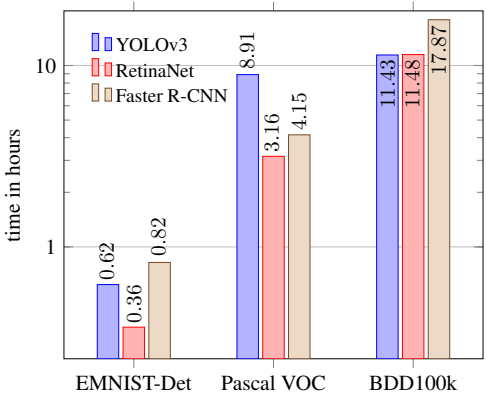
\noindent
AL for advanced image perception tasks such as OD tends to be highly time intensive, compute-heavy and energy consuming.
This is due to the fact that at each AL step the model should be guaranteed to fit to convergence and there are multiple steps of (ideally) several random seeds to be executed.
%\TR{\sout{In order to save development time in AL for OD, we have identified different aspects of complexity in the problem which could be suitably down-scaled to accelerate the process}}.
\cref{fig: runtime} shows the time per AL step used in our setting when run on a Nvidia Tesla V100-SXM2-16GB GPU with a batch size of four.
Note that the time-axis is scaled logarithmically, so the experiments on EMNIST-Det are always faster by at least half an order of magnitude.
The training of YOLOv3 on VOC does not start from COCO-pretrained weights (like YOLOv3+BDD) since the two datasets VOC and COCO are highly similar and VOC consists of sub-classes of the COCO dataset. 
In this case, we opt for ImageNet-pretrained\cite{ILSVRC15} backbone weights just like for the other detectors.
We overall save time up to a factor of around $14$ for VOC and around $32$ for the BDD dataset. %\commentMS{auch mit VOC vergleichen?}
Translated to AL investigations, this means that the effects of new query strategies can be evaluated within half a day on a single Nvidia Tesla V100-SXM2-16GB.

\section{Conclusion}
\label{sec: conclusion}
\noindent
In this work, we investigated the possibility of simplifying the active learning setting in deep object detection in order to accelerate development and evaluation time.
We found that active learning curves do not trivially generalize from one (dataset, detector, evaluation metric)-constellation to another and that significant qualitative differences can be observed.
Even though this prohibits the prediction of final method rankings we observe that the rank correlations are primarily positive and stay high throughout the active learning experiment.
In our evaluation, we also included a more direct measurement of annotation effort in counting the number of queried instances in addition to the number of queried images.
%\TR{\sout{Evaluation of active learning in our laboratory environment leads to behavior that is comparable to standard settings which can be found in the literature.}}
%\HG{Das wirkt widersprüchlich zu [*]. Ich verstehe, dass die Rang-Korrelationen in Fig. 6 nicht viel mehr hergeben. Haben wir dann unseren Ansatz selbst widerlegt? Dann können wir aber auch nicht mehr dafür argumentieren. Meiner meinung nach sollten wir in [*] sagen, dass, obschon die genaue Reihenfolge unvorhersehbar ist, die Korrelationen immerhin durchgängig positiv sind und zT auch recht stark. }
Meanwhile, we can save more than an order of magnitude in total compute time due to the down-scaling of the detection architecture and simplifying the dataset complexity.
Our environment allows for consistent benchmarking of active learning methods in a unified object detection framework, thereby improving transparency.
%\sout{Therefore, development of new methods is highly accelerated, transparent and trackable.} \HG{diese conclusions sind nur richtig, wenn wir an usere methode auch gleuben}
We hope that the present sandbox environment and findings along with the implemented framework will lead to further and accelerated progress in the field of active learning for object detection.

\noindent
\textbf{Acknowledgement.}
The research leading to these results is funded by the German Federal Ministry for Economic Affairs and Climate Action" within the projects ``KI-Absicherung - Safe AI for Automated Driving'', grant no.\  19A19005R and ``KI Delta Learning - Scalable AI for Automated Driving'', grant no.\ 19A19013Q.
We thank the consortia for the successful cooperation.
We gratefully acknowledge financial support by the state Ministry of Economy, Innovation and Energy of Northrhine Westphalia (MWIDE) and the European Fund for Regional Development via the FIS.NRW project BIT-KI, grant no.\ EFRE-0400216. 
The authors gratefully acknowledge the \href{www.gauss-centre.eu}{Gauss Centre for Supercomputing e.V.} for funding this project by providing computing time through the John von Neumann Institute for Computing on the GCS Supercomputer JUWELS at Jülich Supercomputing Centre.
\includegraphics[width=.5\linewidth]{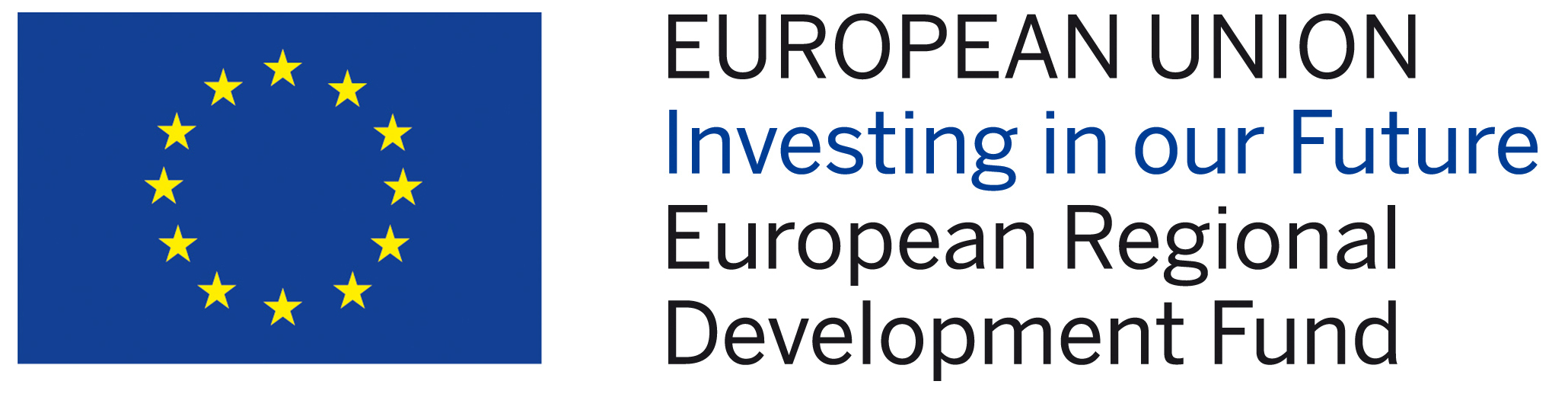}

% \clearpage
\bibliographystyle{ieee_fullname}
\bibliography{biblio}

\clearpage
\appendix
\begin{center}
    Supplementary Material to:
    \\
    \Large
    Towards Rapid Prototyping and Comparability in Active Learning for Deep Object Detection
\end{center}

\section{Implementation Details}
\begin{table}[b]
    \centering
    \caption{Standard deviations of center coordinates, width and height (all relative to image size) of bounding boxes, as well, as number of categories in the training split for several object detection datasets.}
    \resizebox{\linewidth}{!}{
    \begin{tabular}{l|c c c c c}\toprule
        Dataset & $c_x$ & $c_y$ & $w$ & $h$ & \# categories \\
        \midrule
        SVHN & $0.099$ & $0.059$ & $0.048$ & $0.161$ & $10$ \\
        Pascal VOC & $0.217$ & $0.163$ & $0.284$ & $0.277$ & $20$ \\
        MS COCO & $0.254$ & $0.209$ & $0.220$ & $0.234$ & $80$ \\
        KITTI & $0.229$ & $0.080$ & $0.067$ & $0.157$ & $8$ \\
        BDD100k & $0.224$ & $0.133$ & $0.059$ & $0.086$ & $10$ \\
        MNIST-Det & $0.233$ & $0.233$ & $0.054$ & $0.054$ & $10$ \\
        EMNIST-Det & $0.233$ & $0.233$ & $0.066$ & $0.065$ & $26$ \\
        \bottomrule
    \end{tabular}
    }
    \label{tab: dataset varieties}
\end{table}

\subsection{OD Architectures}
\begin{table}
    \centering
    \caption{Configuration of Darknet20 compared with Darknet53 in analogy to \cite[Tab.~ 1]{redmon2018yolov3}.
    At equal resolution input, the feature maps also remain at the same resolution at each stage.
    }
    \resizebox{\linewidth}{!}{
    \input{tables/darknet_tables.tex}
    }
    \label{tab: darknet configs}
\end{table}
In our experiments we used a YOLOv3 detector with the standard Darknet53 backbone on VOC and BDD.
In our down-scaled version we replace the backbone with an analogous architecture working on the same resolutions, such that all strides remain the same and the detection mechanism works identically.
\cref{tab: darknet configs} shows the configuration comparison between the standard Darknet53 architecture and our adapted version (here, called Darknet20) which significantly reduces depth and the number of feature channels extracted.
All other model and data augmentation configurations remain unchanged.

For Faster R-CNN we use a ResNet101 backbone on VOC and BDD while for RetinaNet, we use a ResNet50.
Here, we use a Feature Pyramid Network (FPN) with $[256, 512, 1024, 2048]$ channels.
Both are down-scaled to ResNet18 backbones with $[64, 128, 256, 512]$-channel FPN to accelerate training and inference.

For all architectures, we insert dropout layers between the two last layers (convolutional layers at all stages for YOLOv3/RetinaNet and fully connected for Faster R-CNN).
For all experiments involving sampling, \ie, MC Dropout and Mutual Information experiments, we use dropout rates of $0.5$ and perform $10$ forward passes.

\subsection{Dataset Variability}
\begin{figure*}
    \centering
    \includegraphics[height=0.15\textheight]{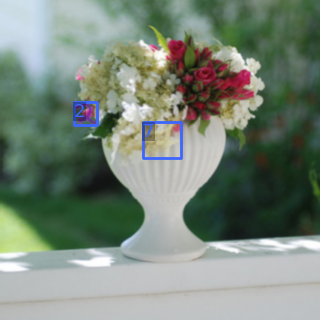}
    \includegraphics[height=0.15\textheight]{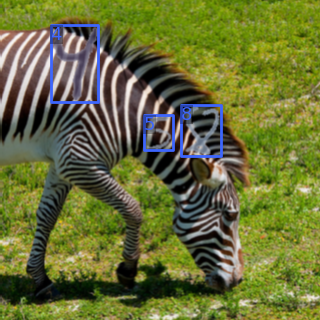}
    \includegraphics[height=0.15\textheight]{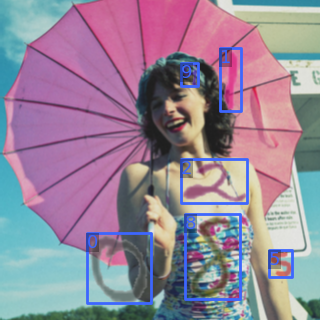}
    \includegraphics[height=0.15\textheight]{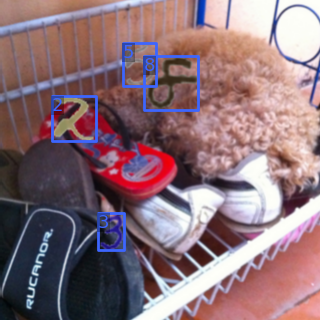}
    \includegraphics[height=0.15\textheight]{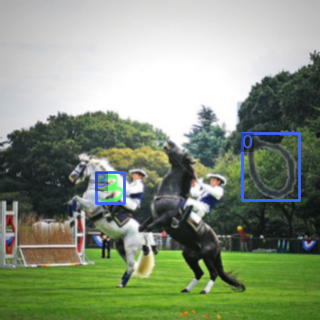}
    \includegraphics[height=0.15\textheight]{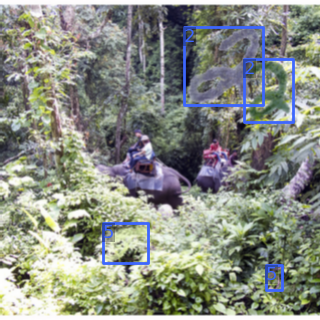}
    \includegraphics[height=0.15\textheight]{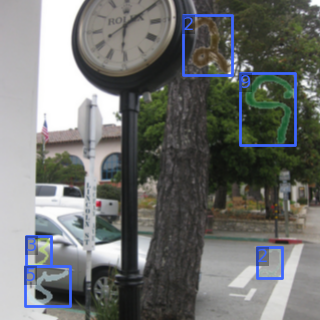}
    \includegraphics[height=0.15\textheight]{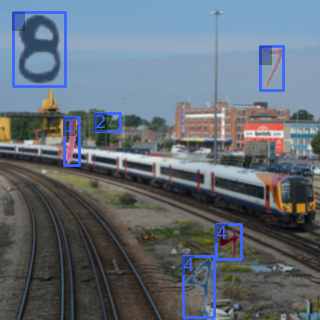}
    \includegraphics[height=0.15\textheight]{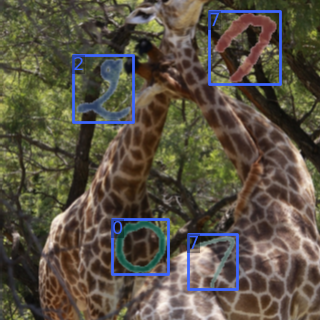}
    \includegraphics[height=0.15\textheight]{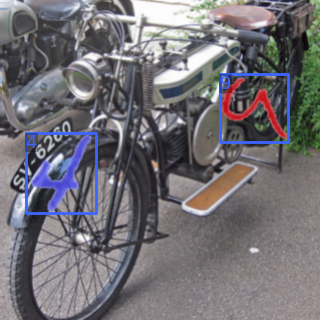}
    \includegraphics[height=0.15\textheight]{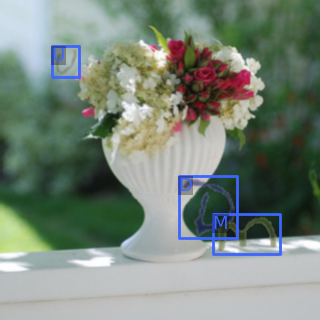}
    \includegraphics[height=0.15\textheight]{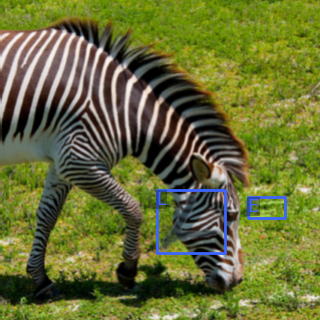}
    \includegraphics[height=0.15\textheight]{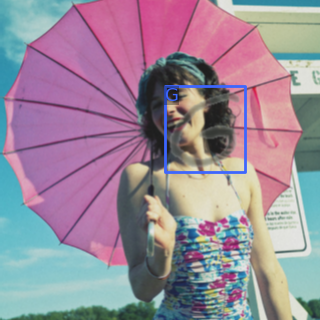}
    \includegraphics[height=0.15\textheight]{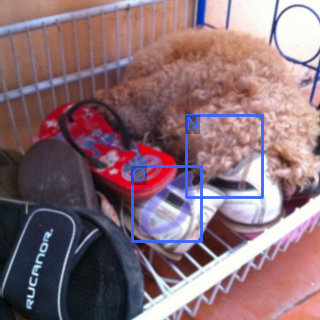}
    \includegraphics[height=0.15\textheight]{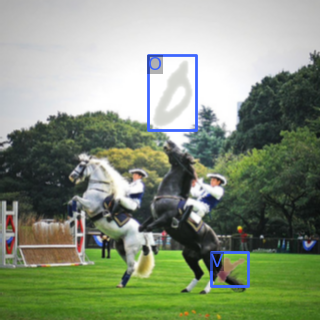}
    \includegraphics[height=0.15\textheight]{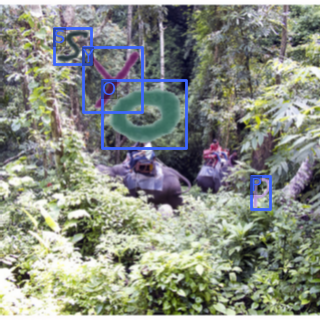}
    \includegraphics[height=0.15\textheight]{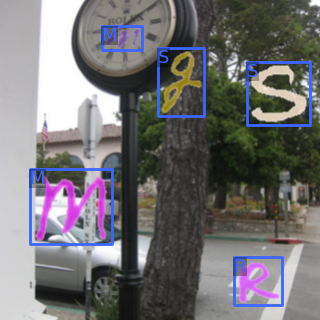}
    \includegraphics[height=0.15\textheight]{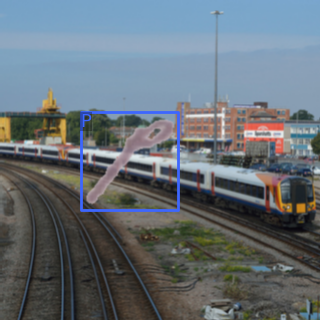}
    \includegraphics[height=0.15\textheight]{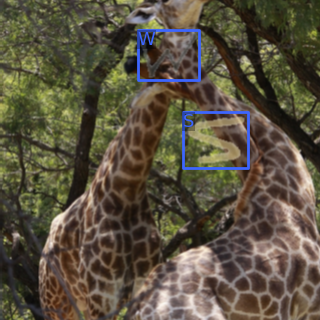}
    \includegraphics[height=0.15\textheight]{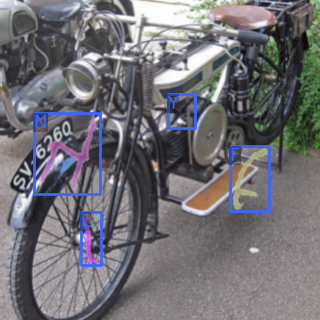}
    \caption{Dataset samples from MNIST-Det (top three rows) and EMNIST-Det (bottom three rows) including annotations.}
    \label{fig: dataset samples}
\end{figure*}

When comparing to existing OD datasets, our sandbox datasets MNIST-Det and EMNIST-Det resemble in variability the common benchmarks like Pascal VOC, MS COCO, KITTI or BDD100k.
This can be seen when looking at the variations in bounding box localization across each dataset.
When normalizing to the total image resolution, we can compare the standard deviations in the annotation center coordinates $c_x$, $c_y$ as well as the bounding box extent $w$ and $h$, which we have collected in \cref{tab: dataset varieties}.
The SVHN dataset consisting of photographs of house numbers shows little variability, especially in the center localization of the object (which are mostly centered on the image).
\Cref{fig: dataset samples} shows samples from the MNIST-Det and EMNIST-Det datasets.

\subsection{AL Parameters}
\begin{table*}
    \centering
    \caption{Overview of important AL hyperparameters for querying data and model training for all datasets and architectures.}
    \resizebox{\linewidth}{!}{
    \begin{tabular}{l l|c c c c c|c c c}\toprule
         &  & \multicolumn{5}{c}{Query} & \multicolumn{3}{c}{Training} \\
         &  & $\vert \mathcal{U}_{init}\vert$ & $\vert\mathcal{Q}\vert$ & $\epsilon_s$ & image resolution & $T$ & batch size & training iters & image resolution \\
         \midrule
         \parbox[t]{0.08\linewidth}{\multirow{4}{*}{\rotatebox[origin=c]{90}{YOLOv3}}} & MNIST-Det & $100$ & $50$ & $0.5$ & $300\times 300$ & $8$ & $4$ & $35,\!000$ & $300\times 300$ \\
         & EMNIST-Det & $100$ & $100$ & $0.5$ & $320\times 320$ & $8$ & $4$ & $50,\!000$ & $320\times 320$ \\
         & Pascal VOC & $200$ & $150$ & $0.5$ & $608\times 608$ & $15$ & $4$ & $18,\!750$\tablefootnote{If the training process for YOLOv3 on VOC is trained with pre-trained ImageNet weights instead of COCO weights, the number of training iterations increases from $18,\!750$ to $200,\!000$.} & $[(320,320),\ (608,608)]$ \\
         & BDD100k & $1,\!100$ & $700$ & $0.5$ & $608\times 608$ & $8$ & $4$ & $160,\!000$ & $[(320,320),\ (608,608)]$\\
         \midrule
         \parbox[t]{0.08\linewidth}{\multirow{4}{*}{\rotatebox[origin=c]{90}{FRCNN}}} & MNIST-Det & $100$ & $50$ & $0.7$ & $300\times 300$ & $8$ & $4$ & $30,\!000$ & $300\times 300$ \\
         & EMNIST-Det & $100$ & $100$ & $0.7$ & $320\times 320$ & $8$ & $4$ & $30,\!000$ & $320\times 320$ \\
         & Pascal VOC & $100$ & $100$ & $0.7$ & $1000\times 600$ & $15$ & $4$ & $18,\!750$ & $1000\times 600$ \\
         & BDD100k & $1,\!250$ & $750$ & $0.7$ & $1000\times 600$ & $8$ & $4$ & $170,\!000$ & $1000\times 600$ \\
         \midrule
         \parbox[t]{0.08\linewidth}{\multirow{4}{*}{\rotatebox[origin=c]{90}{RetinaNet}}} & MNIST-Det & $100$ & $50$ & $0.5$ & $300\times 300$ & $8$ & $4$ & $25,\!000$ & $300\times 300$ \\
         & EMNIST-Det & $225$ & $125$ & $0.5$ & $300\times 300$ & $8$ & $4$ & $35,\!000$ & $300\times 300$ \\
         & Pascal VOC & $550$ & $350$ & $0.5$ & $1000\times 600$ & $8$ & $4$ & $60,\!000$ & $1000\times 600$ \\
         & BDD100k & $1,\!000$ & $500$ & $0.5$ & $1000\times 600$ & $8$ & $4$ & $175,\!000$ & $1000\times 600$ \\
        \bottomrule
    \end{tabular}
    }
    \label{tab: al hyperparameters}
\end{table*}

\cref{tab: al hyperparameters} gives an overview of chosen hyperparameters for the AL cycle for all datasets and architectures. 
Thereby, $\vert \mathcal{U}_{init}\vert$ stands for the number of initially annotated images, $\vert\mathcal{Q}\vert$ for the size of the selected query, $\epsilon_s$ for the score threshold for query inference (thereby, determining instance-wise uncertainty) and $T$ for number of AL steps.
The hyperparameters for training are the batch size, which is always 4, the number of training iterations, and the image resolution.
It follows from \cref{tab: al hyperparameters} (particularly, $\vert \mathcal{U}_{init}\vert$, $\vert\mathcal{Q}\vert$ and $T$) that all architectures considered in our experiments need the fewest images for our sandbox datasets MNIST-Det and EMNIST-Det to reach the $90\%$ max performance mark. 
Therefore, for the sandbox datasets we chose $\vert \mathcal{U}_{init}\vert$ and $\vert\mathcal{Q}\vert$ smaller than for VOC and BDD. 
Moreover, the sandbox datasets have lower image resolutions, which leads to faster training and inference times, even if occasionally the training iterations are lower for VOC, \eg, for Faster R-CNN and YOLOv3.
% \footnote*{${*}$If the training process for YOLOv3 on VOC is trained with pre-trained ImageNet weights instead of COCO weights, the number of training iterations increases from $18,750$ to $200,000$.}
Apart from the latter case, the most iterations to obtain convergence in the training processes are needed for BDD with up to $175,\!000$.
The score threshold of $0.5$ for YOLOv3 and RetinaNet, and $0.7$ for Faster R-CNN was determined by ablation studies for EMNIST-Det and then adopted for the other datasets.
For all query methods, we incorporate a class-weighting (the same as in \cite{brust2018active}) for computing instance-wise uncertainty scores.

\section{Ablations}
\subsection{Image-aggregation Methods}
\begin{figure*}
    \centering
    \includegraphics[height=0.23\textheight]{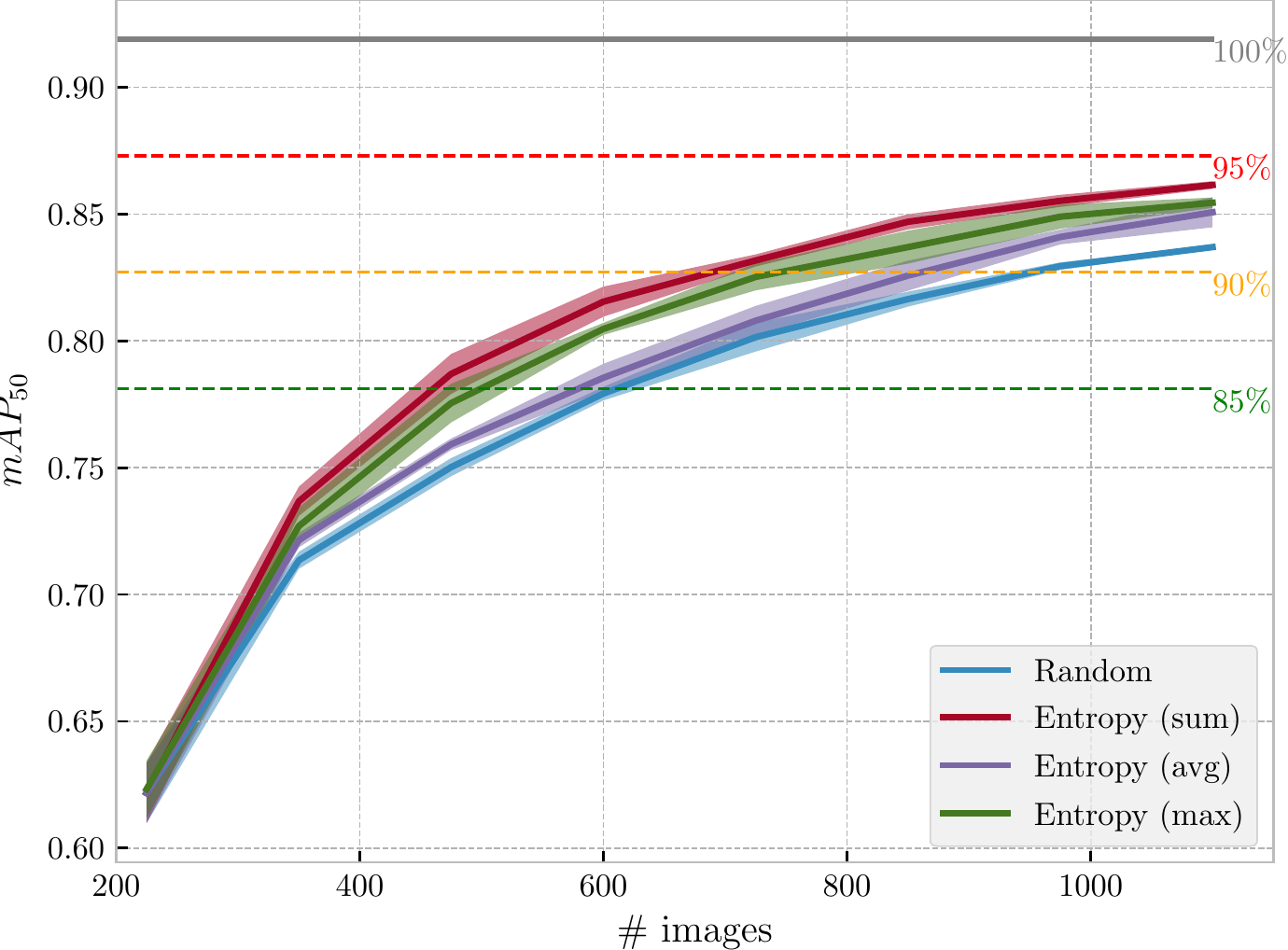}
    \includegraphics[height=0.23\textheight]{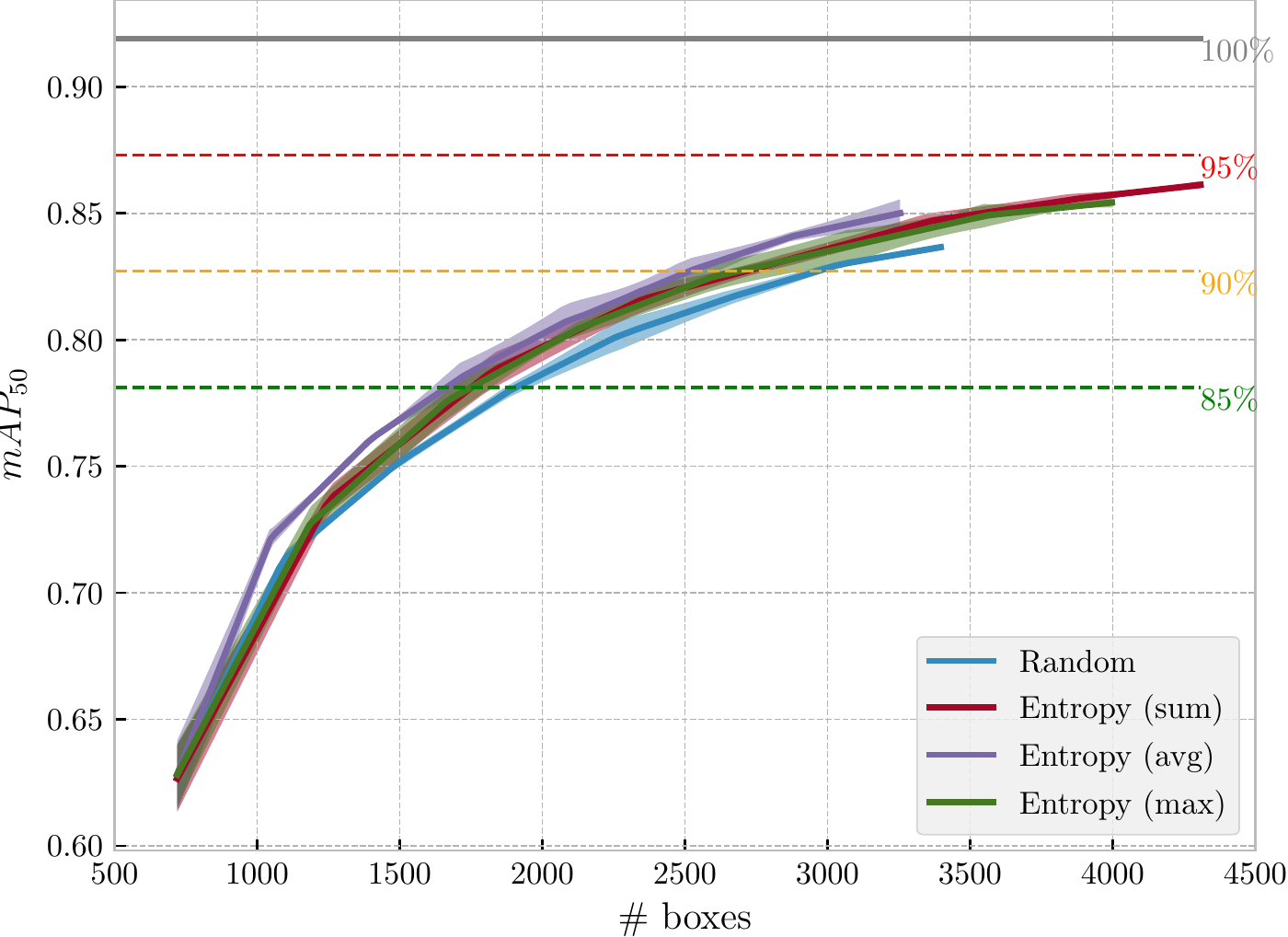}
    \includegraphics[height=0.23\textheight]{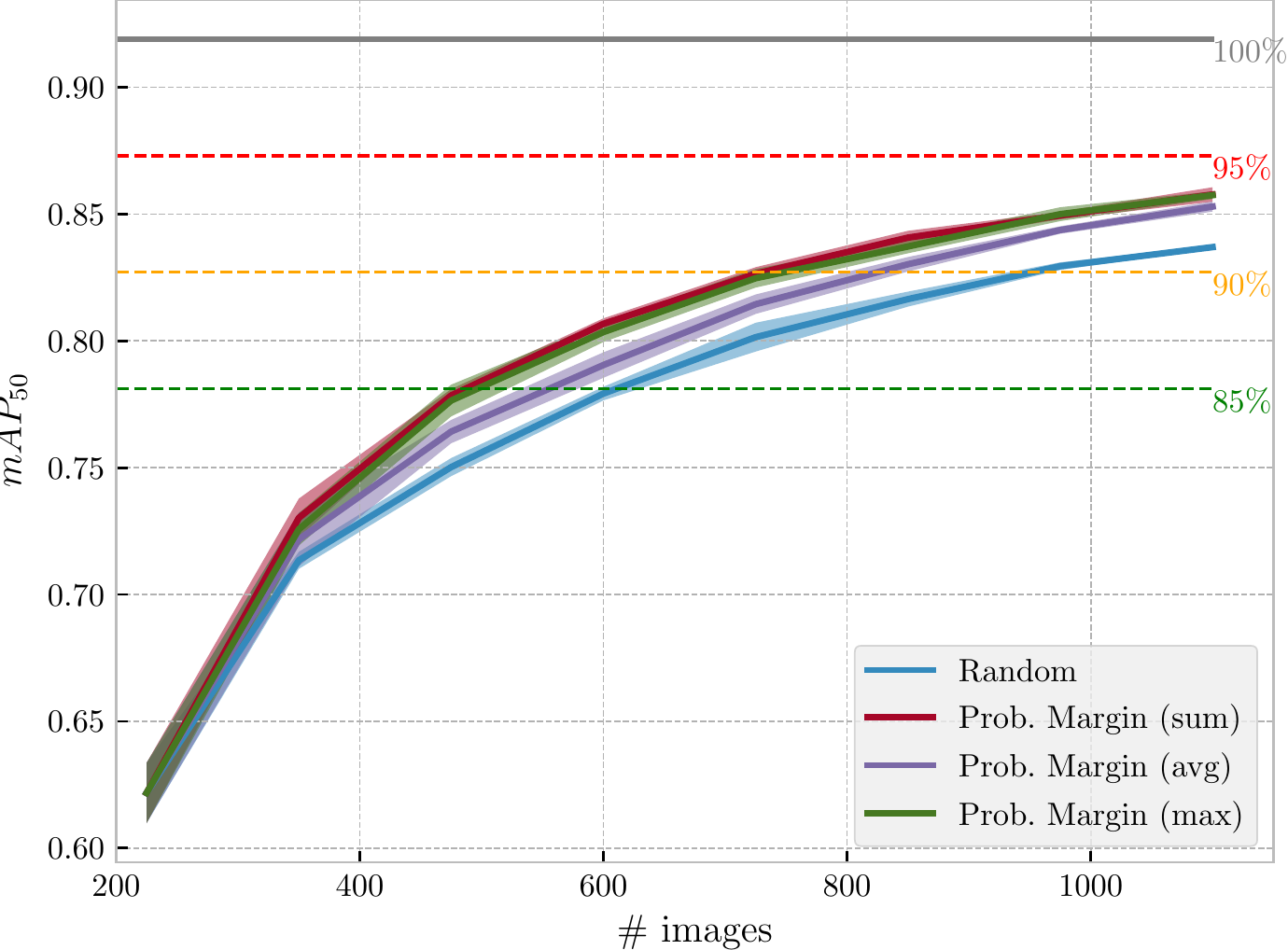}
    \includegraphics[height=0.23\textheight]{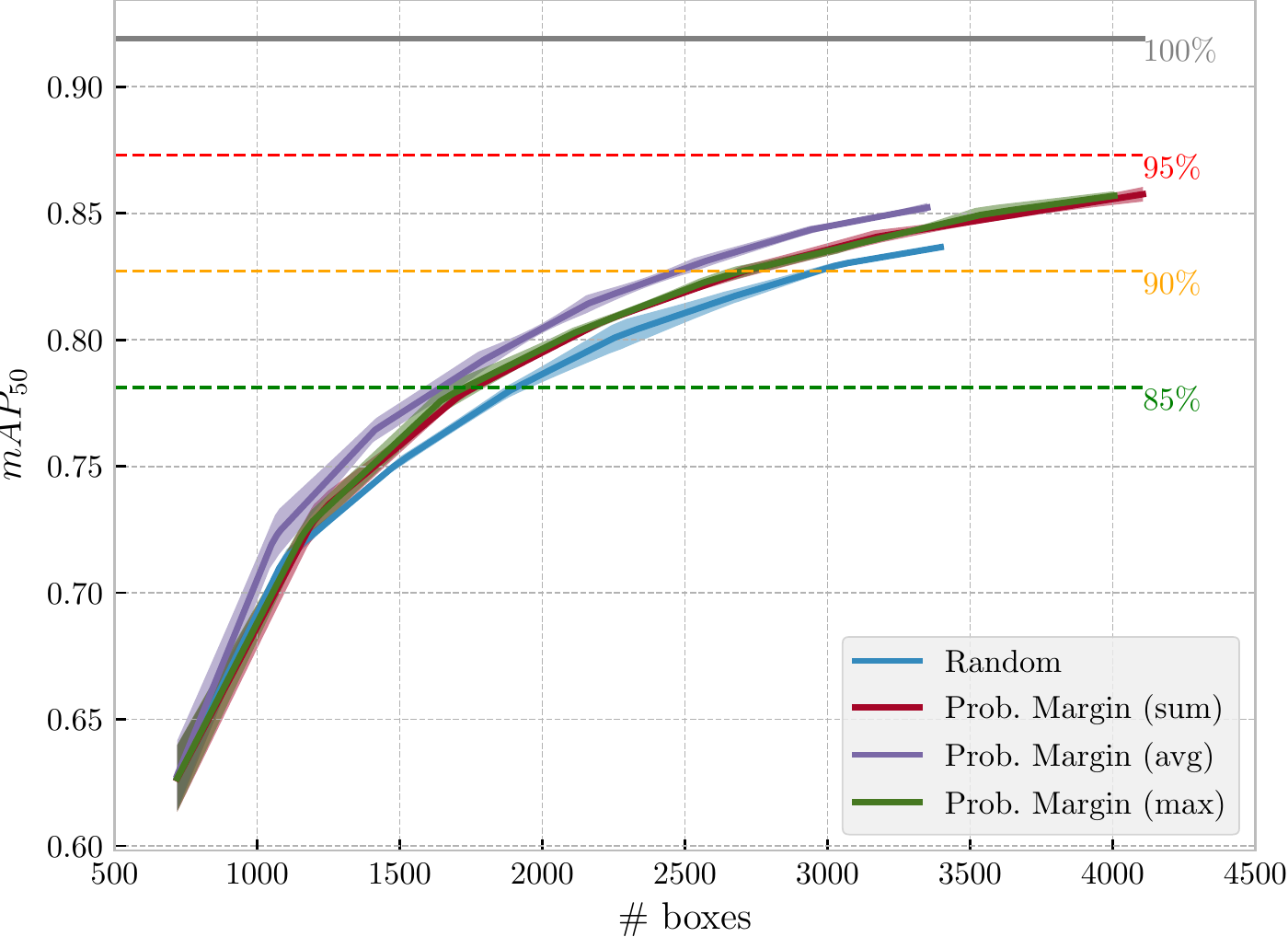}
    \includegraphics[height=0.23\textheight]{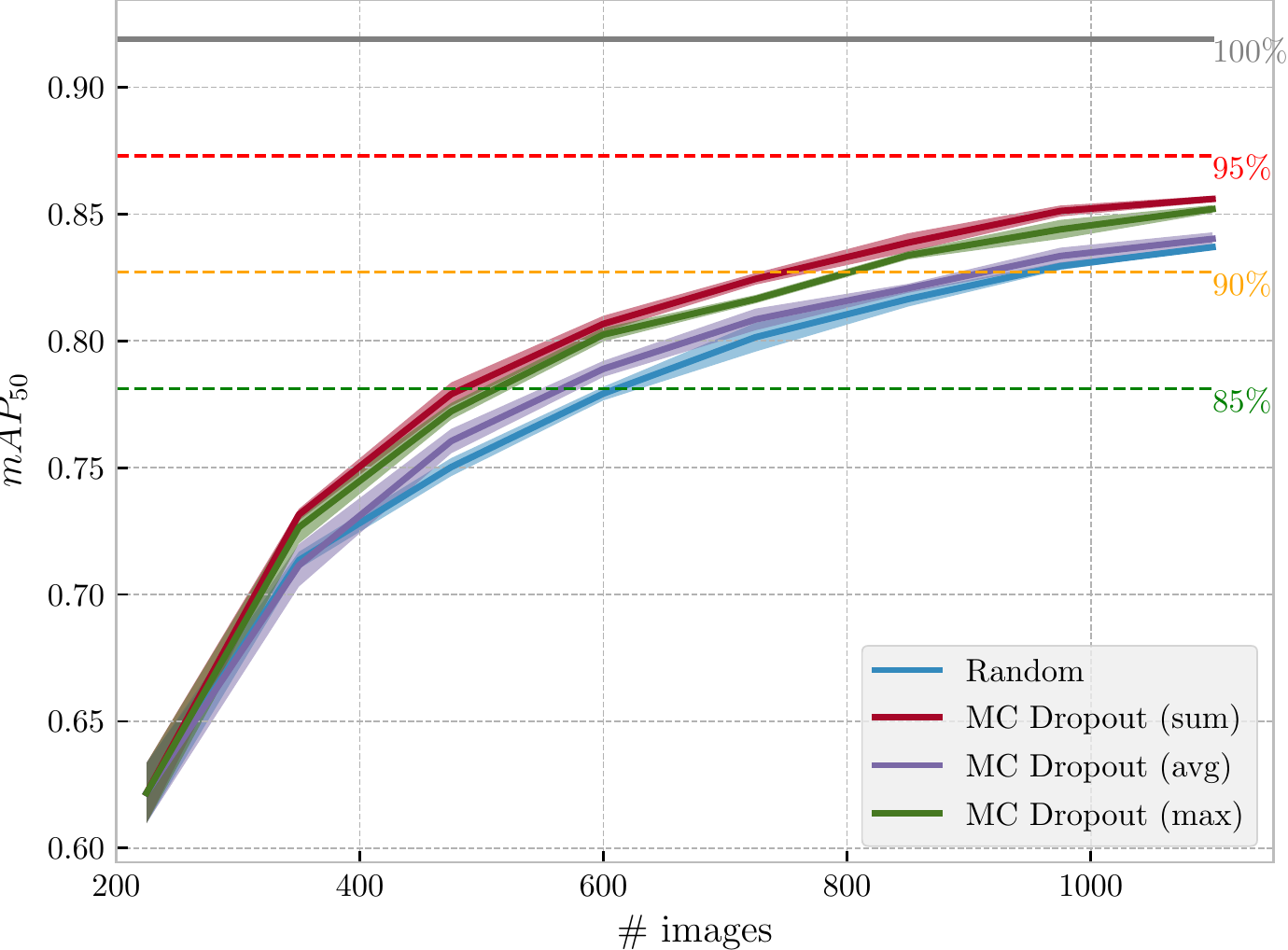}
    \includegraphics[height=0.23\textheight]{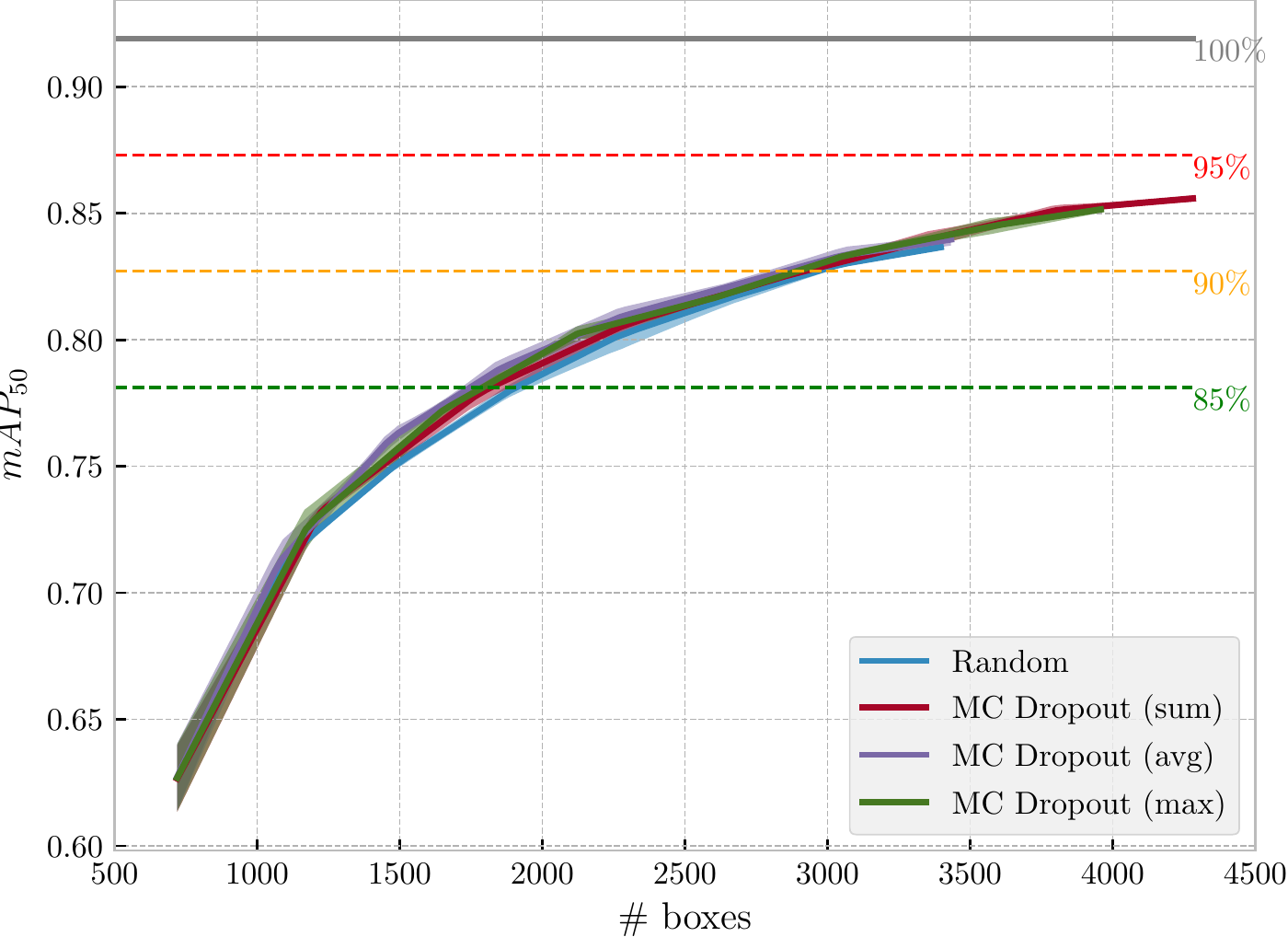}
    \includegraphics[height=0.23\textheight]{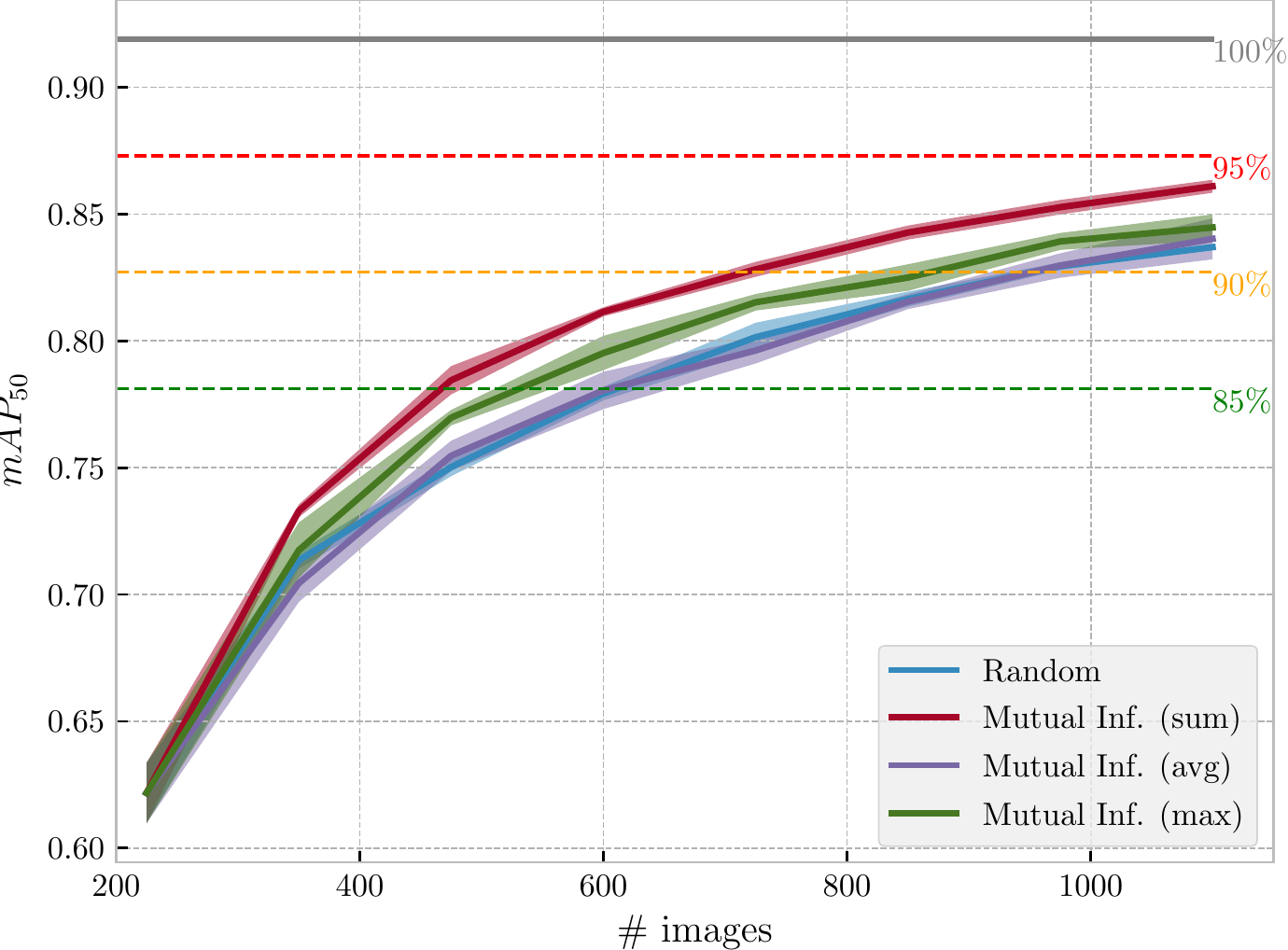}
    \includegraphics[height=0.23\textheight]{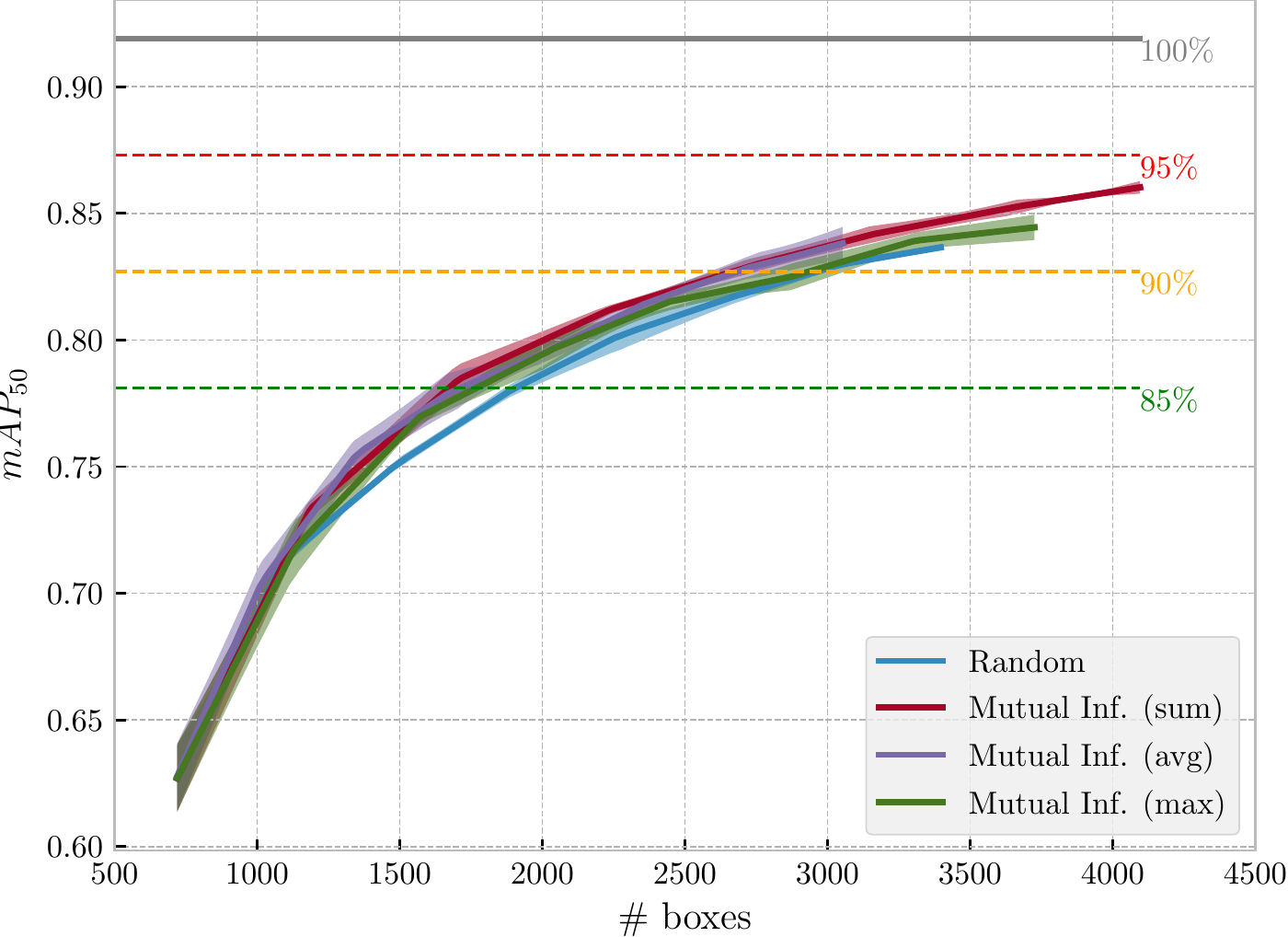}
    \caption{Ablation study on the aggregation method for RetinaNet on the EMNIST-Det dataset.}
    \label{fig: retinanet aggregations}
\end{figure*}

\Cref{fig: retinanet aggregations} shows test $\map_{50}$ for different image aggregations, namely sum, average and maximum, for the RetinaNet on EMNIST-Det. 
The left panels show $\map_{50}$ scores as a function of the number of queried images while the right panels show $\map_{50}$ scores as a function of the number of queried instances. 
For all four uncertainty baselines, the sum dominates the maximum and the maximum dominates the average in the image-wise evaluation. 
Nevertheless, the average remains consistently better than Random, with the exception of Mutual Information, where both curves are almost on par.
A clearly different course is obtained when considering the instance-wise evaluation.
For the same number of images queried, the sum prefers images with many boxes, whereas the average queries images with even fewer boxes than the Random baseline. 
In terms of performance, the average outperforms the sum and maximum, which are tied, and the Random baseline for Entropy and Probability Margin. 
For Mutual Information, the average and sum are best, whereas for MC Dropout all curves are hardly distinguishable from each other.
Comparable behaviors could also be observed on the other architectures and datasets.

Investigations of the kind presented here under normal conditions (using a full-scale standard object detector and a benchmark dataset) would require weeks of compute time and yield valuable information on sensitive parameters for querying.
Using our sandbox environment makes extensive ablation studies of hyperparameters possible within a few days.

\section{Additional Numerical Results}
In this section, we compile additional experimental results supplementing \cref{tab: query amounts} from the main manuscript with additional metrics and results from the RetinaNet detector which were struck from the main part due to spacial constraints.

\begin{table}
    \centering
    \caption{Mean average precision results per query method for maximal amount of images selected; higher values are better. 
    Bold numbers indicate the highest performance per experiment and underlined numbers are the second highest.}
    \resizebox{\linewidth}{!}{\input{tables/map_table.tex}}
    \label{tab: map table}
\end{table}

\begin{table*}
    \centering
    \caption{Area under AL curve results per query method for maximal amount of data (images/bounding boxes) selected; higher values are better. 
    Bold numbers indicate the highest $\auc$ per experiment and underlined numbers are the second highest.}
    \resizebox{\textwidth}{!}{
    \input{tables/auc_table.tex}
    }
    \label{tab: auc table}
\end{table*}

\begin{table*}
    \centering
    \caption{Amount of queried images and bounding boxes necessary to cross the 90\% performance mark during AL for the RetinaNet detector. Lower values are better. Bold numbers indicate the lowest amount of data per experiment and underlined numbers are the second lowest.}
    \resizebox{\textwidth}{!}{
    \input{tables/retinanet_table.tex}
    }
    \label{tab: queried data retinanet}
\end{table*}

\subsection{Benchmark Results}
\Cref{tab: map table} shows the  $\map_{50}$ achieved after the final query for each method and detector-dataset constellation in the style of \cref{tab: query amounts}.
For each AL curve, the final performance (most queried images allowed according to \cref{tab: al hyperparameters}) is independent of an image- or instance-wise point of view.
Overall, the Entropy baseline is consistently among the best two methods, however, the overall rankings show a medium degree of variance across datasets and across detectors especially when regarding instance-wise evaluation in \cref{tab: query amounts}.
Comparing vertical sections through AL curves shows overall roughly similar behavior as the results in \cref{tab: query amounts} (horizontal sections), however, we observe differences in the method rankings in terms of amount of data queried vs.\ final detection performance (\eg, Faster R-CNN on the MNIST-Det dataset).

In \cref{tab: auc table} we collect the values of the $\auc$ metric.
Note, that the $\auc$ metric scales with the $t$-axis, \ie, results between different datasets can only be compared qualitatively.
The same is true for comparisons between image- and instance-wise evaluations.
When comparing with \cref{tab: query amounts} from \cref{sec: benchmark results}, we see a high degree of ranking similarity with the amount of data required to cross the $90\%$-mark in both, image- and instance-wise evaluation.
We conclude with previous findings on the rank correlations, that even in a rather late evaluation (when some fixed reference mark in performance has already been crossed), the $\auc$ shows more similarity with the $90\%$ ranking than raw detection performance (\cref{tab: map table}).
For instance, compare the Faster R-CNN row from \cref{tab: query amounts} with the corresponding row in \cref{tab: auc table}.

We include results for the RetinaNet detection analogous to \cref{tab: query amounts} in \cref{tab: queried data retinanet}.
Again, we notice striking ranking differences between image- and instance-wise evaluation.
Particularly, instance-wise evaluation has rather irregular method rankings between datasets for the RetinaNet detector which were, however, also reflected in our $\auc$ evaluation discussed before in \cref{tab: auc table}. These results yield further evidence that the consideration of only a single evaluation metric for active learning performance is insufficient.

\subsection{Generalization on RetinaNet}

\begin{figure}
    \centering
    \rotatebox{90}{\qquad\quad$\map_{50}$}
    \includegraphics[height=3cm]{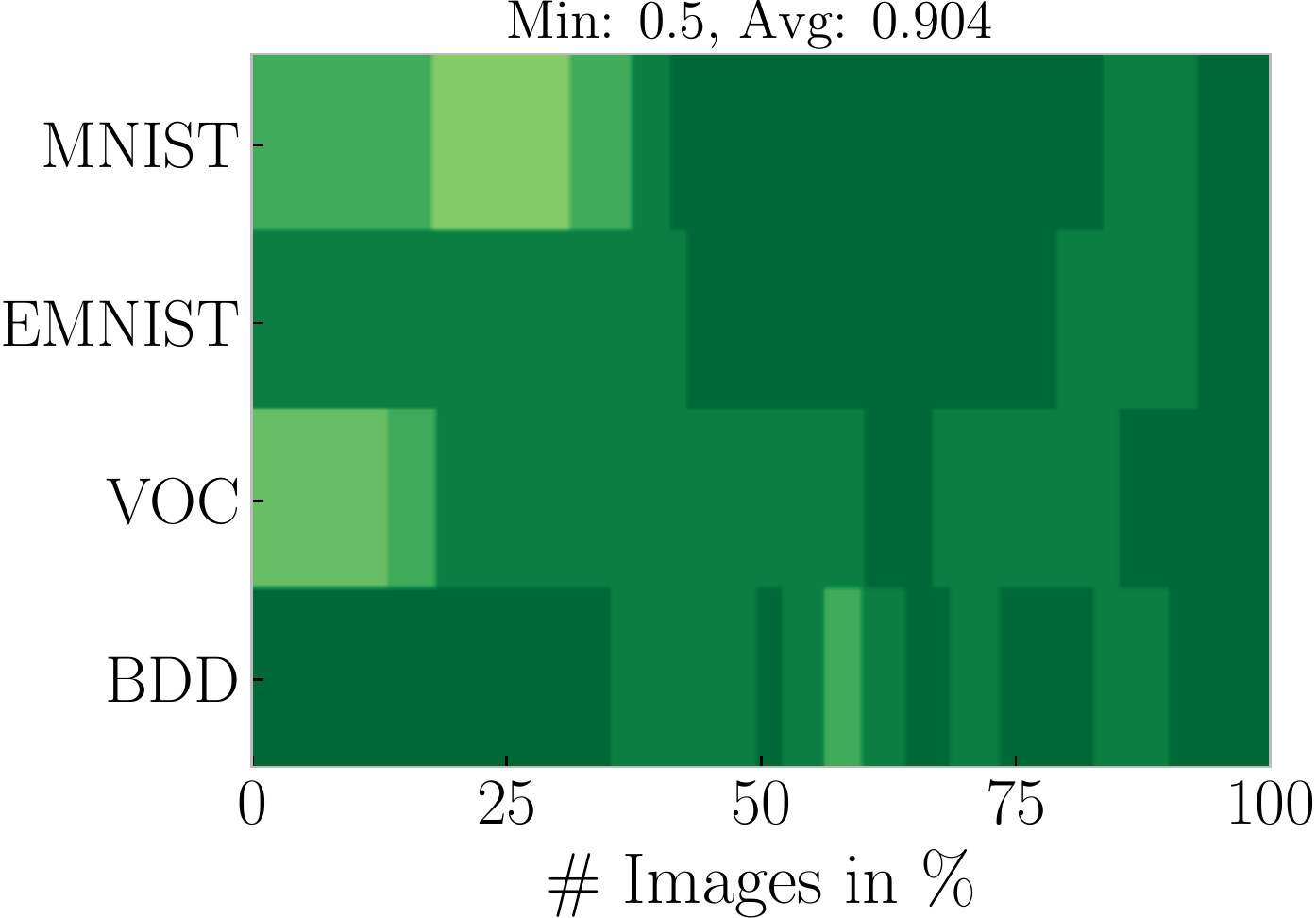}
    \includegraphics[height=3cm]{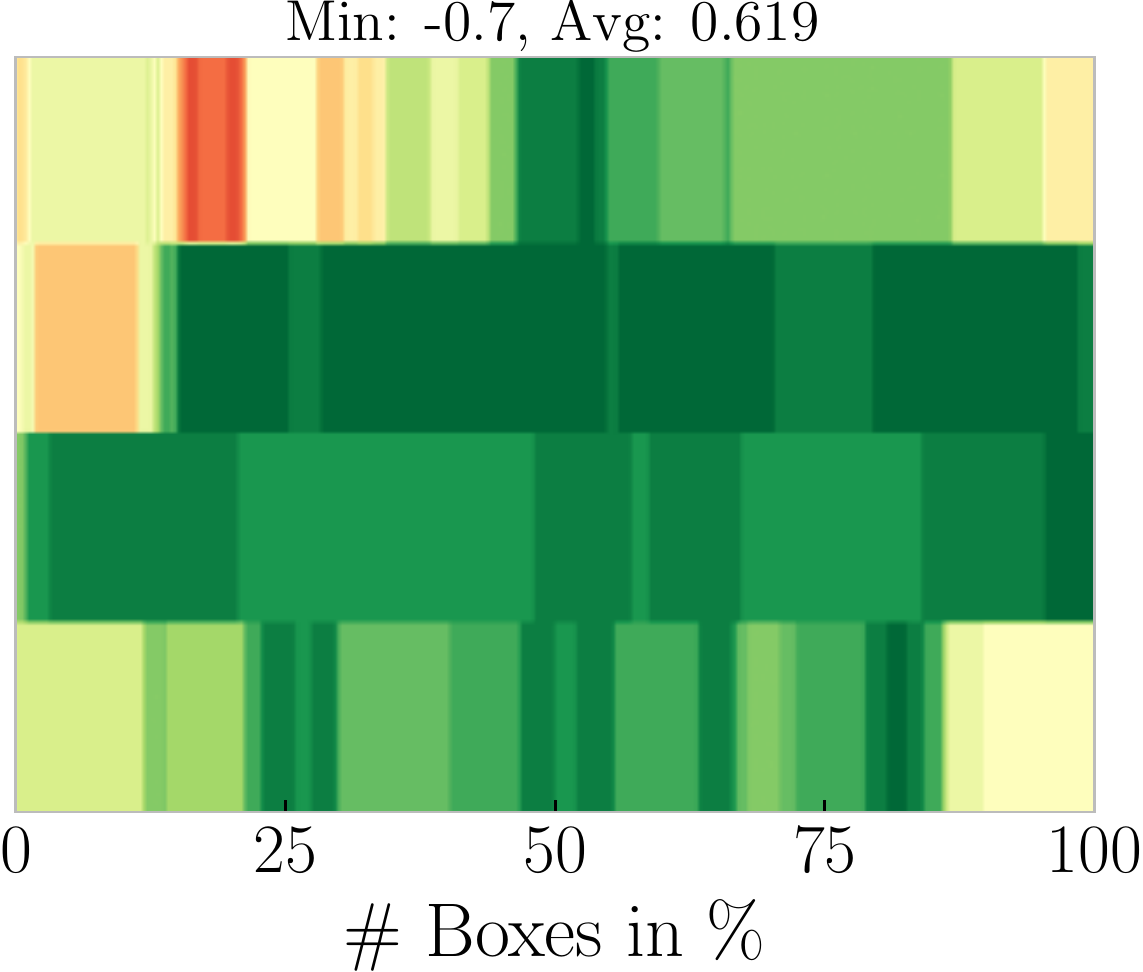}
    \vspace{0.0cm}
    \newline
    \rotatebox{90}{\qquad\quad$\auc$}
    \includegraphics[height=3cm]{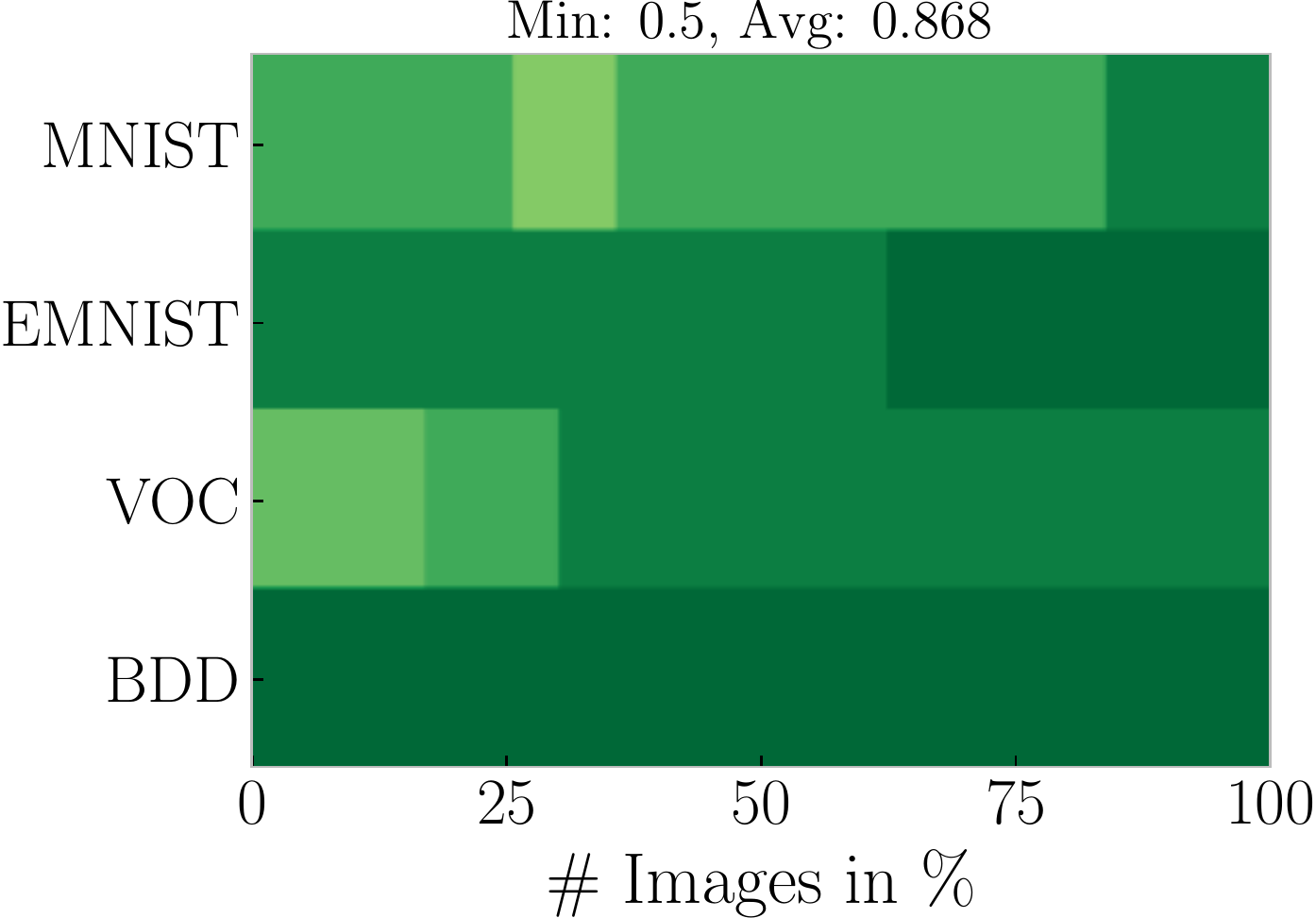}
    \includegraphics[height=3cm]{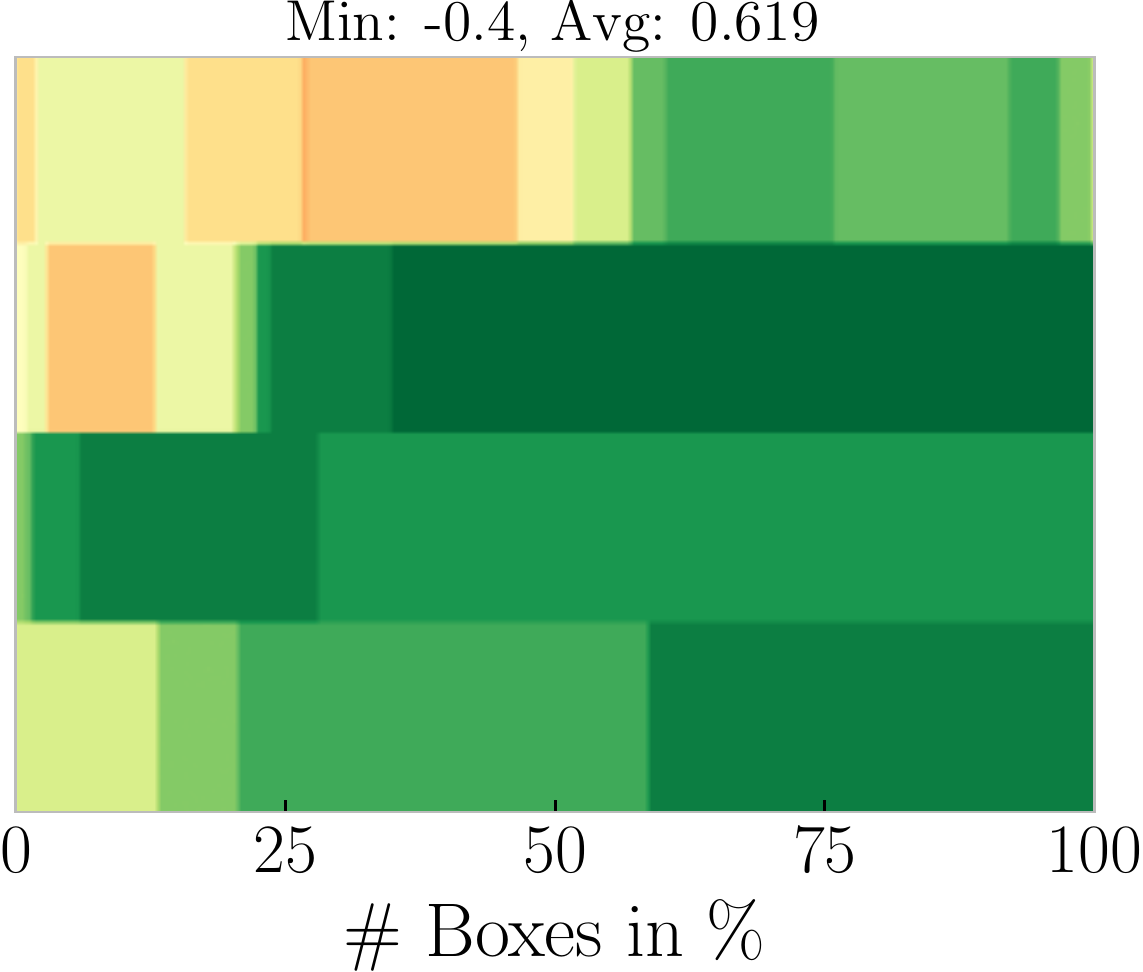}
    \caption{Curves of rank correlations between the cumulative $\auc$ and the final rankings at the $90\%$ max performance mark for the RetinaNet detector.}
    \label{fig: correlation histories retinanet}
\end{figure}

\begin{figure}
    \centering
    \includegraphics[height=3.625cm]{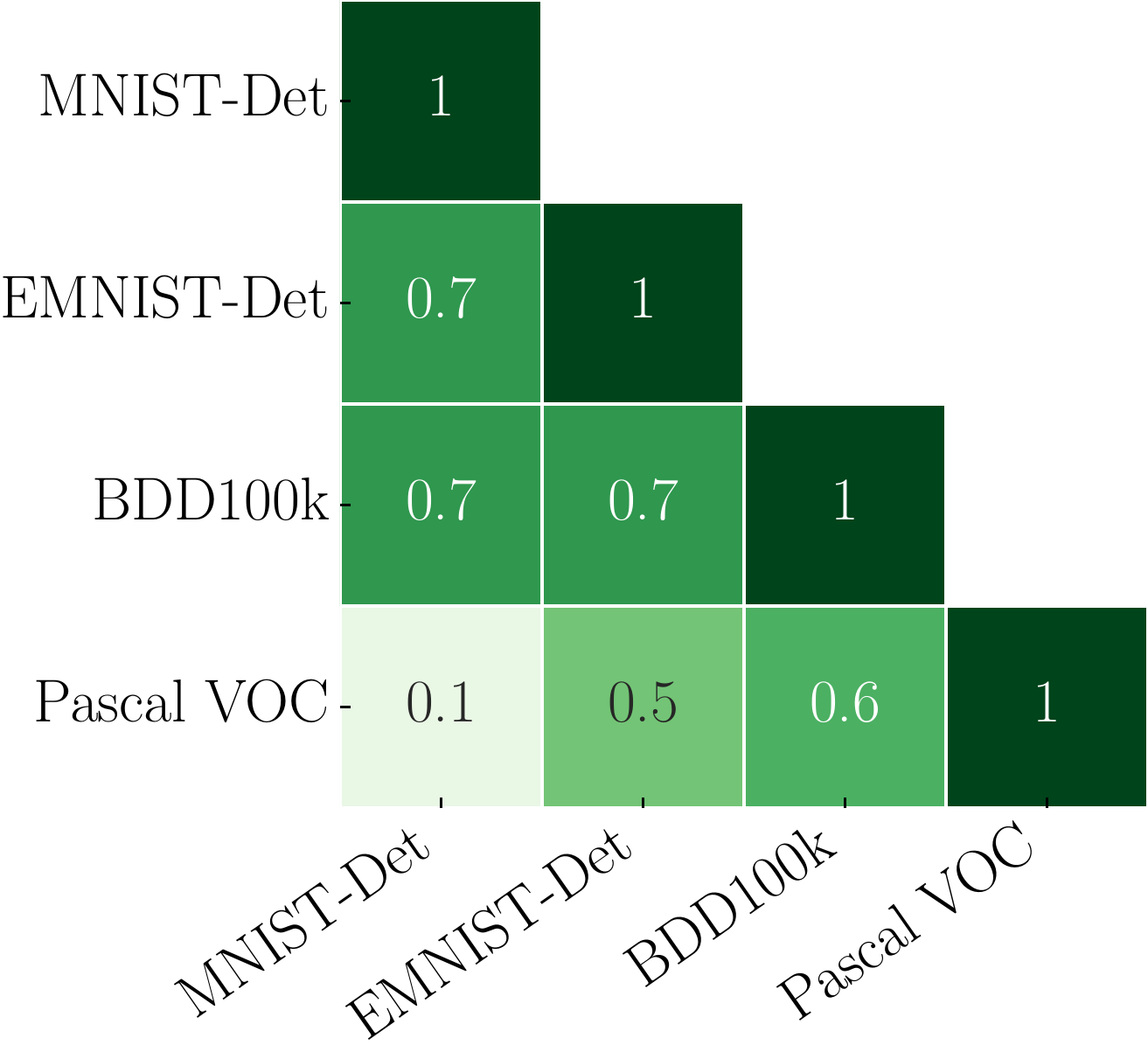}
    \includegraphics[height=3.625cm]{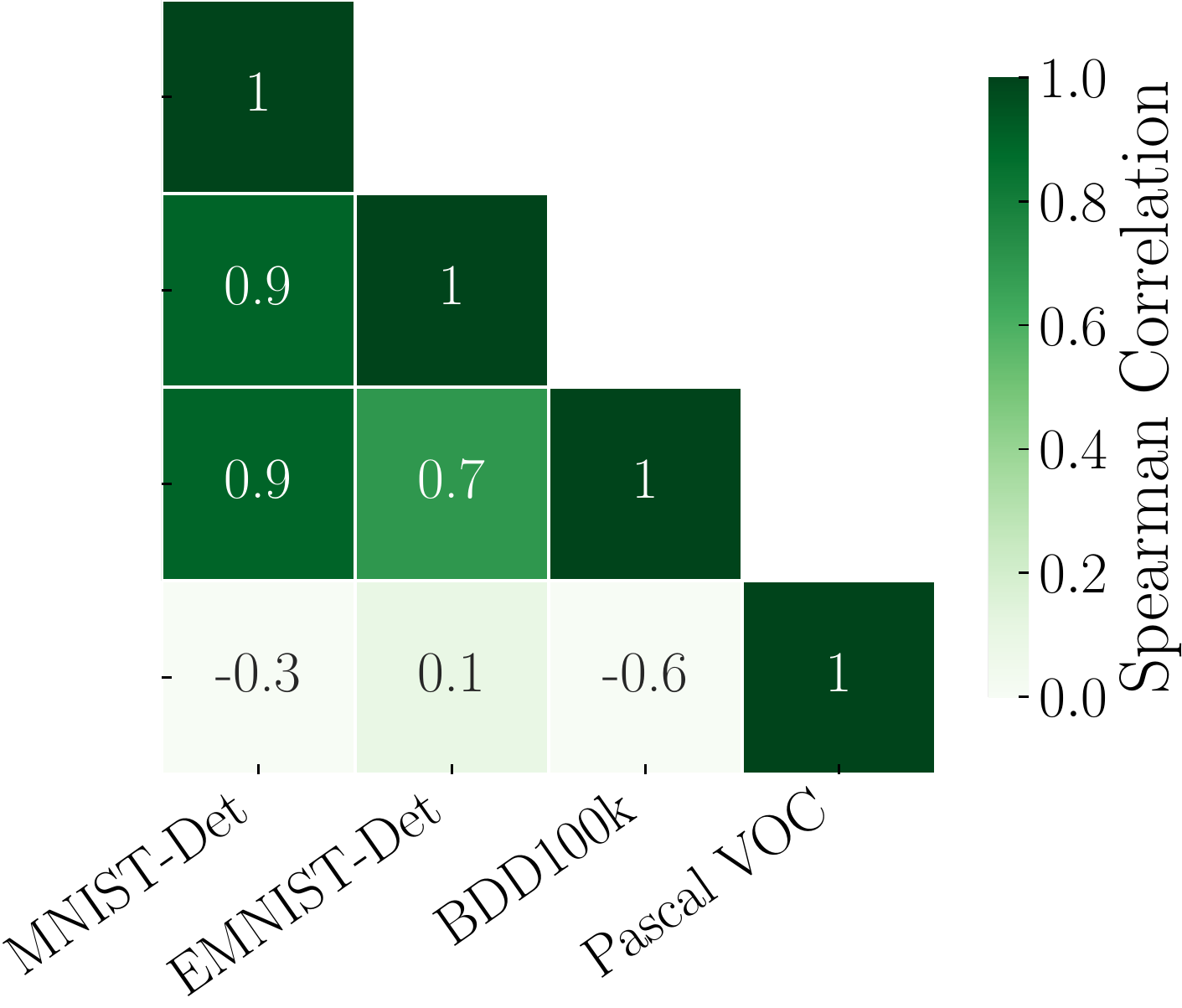}
    \caption{Ranking correlations between $\auc$ values for the RetinaNet detector, left: image-wise; right: instance-wise evaluation.}
    \label{fig: correlation matrices retinanet}
\end{figure}
\begin{figure*}[b]
    \centering
    \def\figheight{3.3cm}
    \includegraphics[height=\figheight]{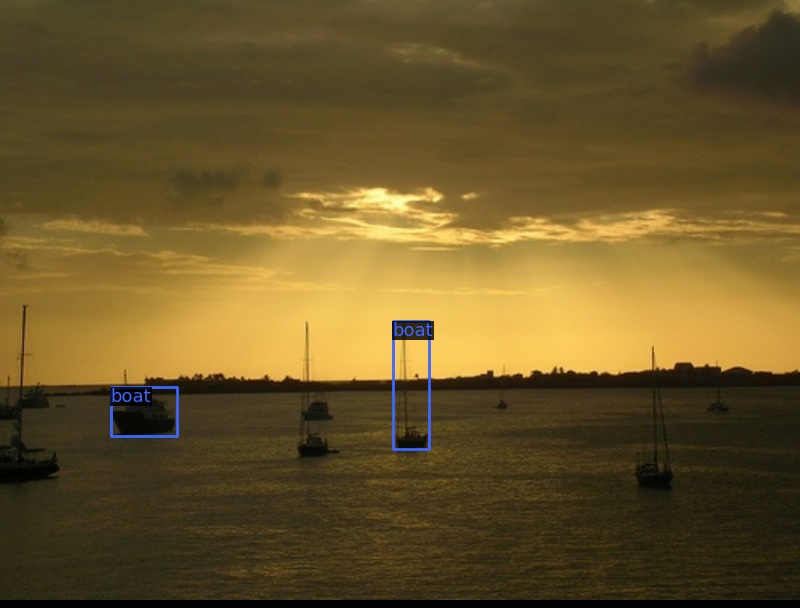}
    \includegraphics[height=\figheight]{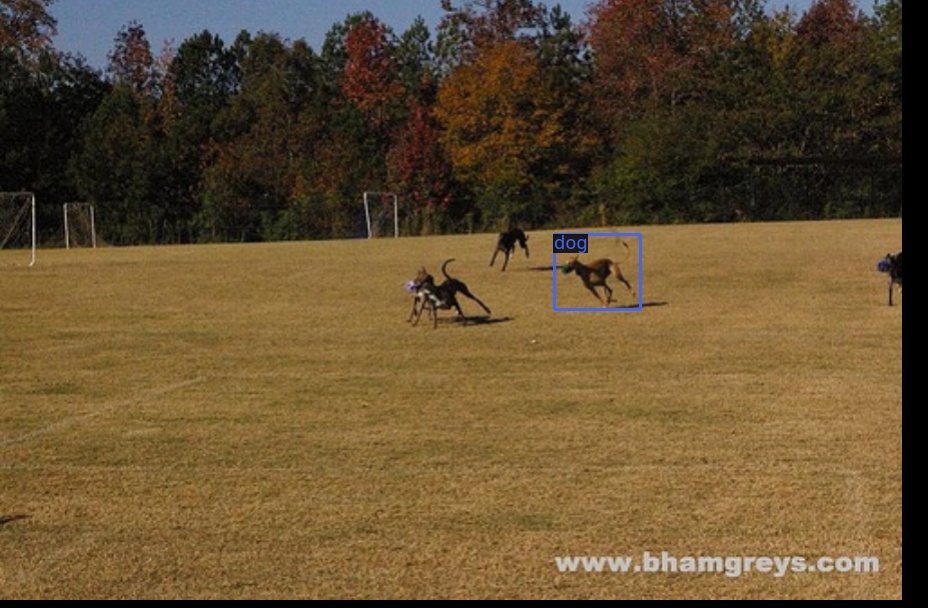}
    \includegraphics[height=\figheight]{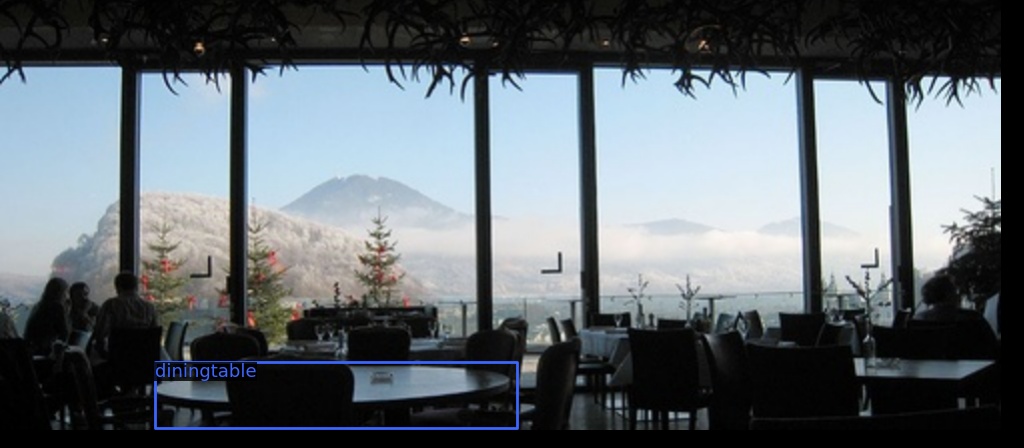}
    \caption{Annotation examples from the Pascal VOC test dataset.}
    \label{fig: voc label errors}
\end{figure*}

Ranking correlations for $\map_{50}$ and $\auc$ for RetinaNet, see \cref{fig: correlation histories retinanet}, tend to show the same behavior as for the YOLOv3 and Faster R-CNN. 
The $\auc$ fluctuates less then the $\map_{50}$ and both metrics show overall high correlation with the $90\%$ max performance ranking.
For the image-wise evaluation, both metrics have correlations greater or equal than $0.5$.
In the instance-wise evaluation, on the other hand, the correlations increase only gradually. 
Even though the average correlations are identical for the $\map_{50}$ and the $\auc$, the latter is clearly more stable w.r.t. the $90\%$ max performance ranking and has a higher minimum correlation.

\Cref{fig: correlation matrices retinanet} shows correlation matrices for the RetinaNet. 
The image-wise comparison shows the highest correlation of $0.7$ when comparing MNIST-Det, EMNIST-Det and BDD respectively. 
VOC has the highest correlation with BDD of $0.6$, but the correlation with EMNIST-Det is similar with $0.5$. 
No conclusions can be drawn between the rankings of MNIST-Det and VOC due to the low correlation of $0.1$.
In the instance-wise comparison, MNIST-Det has very high correlations of $0.9$ with EMNIST-Det and BDD, and the comparability of the rankings of EMNIST-Det and BDD is also given by a correlation of $0.7$.
However, it is again noticeable that VOC is hardly comparable with any other dataset. 
This could be attributed to some of the dataset characteristics. 
On one hand, we observed many missing labels when looking at the VOC data (\cf \cref{fig: voc label errors}). 
On the other hand, the instance sizes of BDD and EMNIST-Det/MNIST-Det seem to be rather comparable as opposed to the instance sizes in VOC.

\end{document}

%% file: definitions.tex
\newcommand{\auc}{\mathit{AUC}}
\newcommand{\map}{\mathit{mAP}}

%% file: figures/cycle.tex
% \documentclass[border=0.2cm]{standalone}
 
% \usepackage{amsmath}
% \usepackage{graphicx}
% \usepackage{pgfplots}
% \usepackage{pgf-pie}  
% \usepackage{tikz}
% \usetikzlibrary{shapes.geometric,arrows.meta,bending,positioning,patterns}
 
% \begin{document}
 
\begin{tikzpicture}
    \node[cylinder, 
        draw=blue,
        cylinder uses custom fill,
        cylinder body fill=blue!20,
        cylinder end fill=blue!40,
        shape border rotate=90,
        aspect=0.7,
        minimum width=0.8cm,
        minimum height=1cm,
        ] (L) at (-1, 0) {$\mathcal{L}$};

    % \node (dnn) at (2, 2) {DNN};
    \begin{scope}[local bounding box=dnn,
        shift={(2,1.5)}]
        \node (00) at (-0.8, -0.4) {};
        \node (01) at (-0.8, 0.0) {};
        \node (02) at (-0.8, 0.4) {};
        \draw[black] (00) circle (2pt);
        \draw[black] (01) circle (2pt) node{};
        \draw[black] (02) circle (2pt) node{};

        \node (10) at (-0.3, -0.6) {};
        \node (11) at (-0.3, -0.2) {};
        \node (12) at (-0.3, 0.2) {};
        \node (13) at (-0.3, 0.6) {};
        \draw[black] (10) circle (2pt) node{};
        \draw[black] (11) circle (2pt) node{};
        \draw[black] (12) circle (2pt) node{};
        \draw[black] (13) circle (2pt) node{};

            \draw[-] (00) -- (10);
            \draw[-] (00) -- (11);
            \draw[-] (00) -- (12);
            \draw[-] (00) -- (13);
            \draw[-] (01) -- (10);
            \draw[-] (01) -- (11);
            \draw[-] (01) -- (12);
            \draw[-] (01) -- (13);
            \draw[-] (02) -- (10);
            \draw[-] (02) -- (11);
            \draw[-] (02) -- (12);
            \draw[-] (02) -- (13);

        \node (20) at (0.2, -0.4) {};
        \node (21) at (0.2, 0.0) {};
        \node (22) at (0.2, 0.4) {};
        \draw[black] (20) circle (2pt) node{};
        \draw[black] (21) circle (2pt) node{};
        \draw[black] (22) circle (2pt) node{};

            \draw[-] (10) -- (20);
            \draw[-] (10) -- (21);
            \draw[-] (10) -- (22);
            \draw[-] (11) -- (20);
            \draw[-] (11) -- (21);
            \draw[-] (11) -- (22);
            \draw[-] (12) -- (20);
            \draw[-] (12) -- (21);
            \draw[-] (12) -- (22);
            \draw[-] (13) -- (20);
            \draw[-] (13) -- (21);
            \draw[-] (13) -- (22);

        \node (30) at (0.7,0.2) {};
        \node (31) at (0.7,-0.2) {};
        \draw[black] (30) circle (2pt) node{};
        \draw[black] (31) circle (2pt) node{};

            \draw[-] (20) -- (30);
            \draw[-] (20) -- (31);
            \draw[-] (21) -- (30);
            \draw[-] (21) -- (31);
            \draw[-] (22) -- (30);
            \draw[-] (22) -- (31);
    \end{scope}
    \node (dnn text) at (2,2.5) {object detector};
    \draw[-Latex,out=80,in=180,draw=blue] (L) to node[midway,above] {\color{blue} {\Large\faHourglassHalf} training} (dnn);
    
    % \node at (1,2) {\color{blue}{\Large\faHourglassHalf}};

    \node[cylinder, 
        draw=gray,
        cylinder uses custom fill,
        cylinder body fill=gray!20,
        cylinder end fill=gray!40,
        pattern=north west lines, 
        pattern color=purple!40,
        shape border rotate=90,
        aspect=0.7,
        minimum width=1.5cm,
        minimum height=1cm
        ] (U) at (5.4, -0.9) {$\mathcal{U}$};

    \node[cylinder, 
        draw=purple,
        cylinder uses custom fill,
        cylinder body fill=purple!20,
        cylinder end fill=purple!40,
        shape border rotate=90,
        aspect=0.4,
        ] (Q) at (2.5, -1.5) {$\mathcal{Q}$};

    \draw[-Latex,out=320,in=55,draw=purple] (dnn) to (Q);
    \draw[-Latex,out=170,in=55,draw=purple] (U) to node[midway,below] {\color{purple} query \small \faHourglassHalf} (Q);

    \draw[-Latex,out=185,in=280] (Q) to node[midway,below] {labeling by oracle} (L);
    \node[draw,ellipse] at (1.2, 0.1) {$t = 1, \ldots, T$};
    \node (oracle) at (0.6,-0.75) {\includegraphics[width=1.5cm]{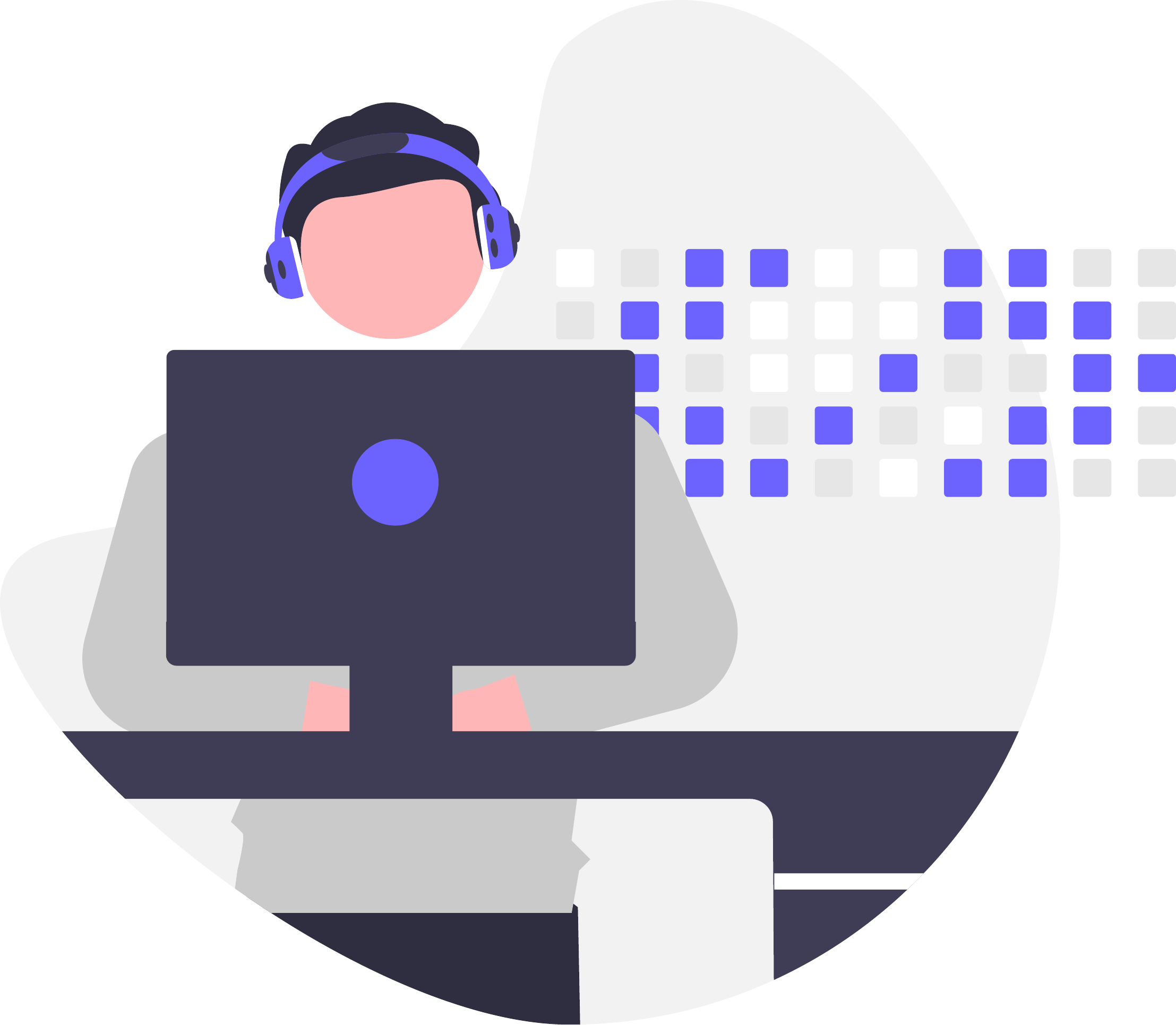}};

    \begin{scope}[local bounding box=perfplot,
            shift={(4.3,0.8)},
            scale=0.5
            ]
        \begin{axis}[
            width=5cm,
            ticks=none,
            xlabel=$t$
            ]
            \addplot[color=blue,mark=o,mark options={fill=black,scale=1}] coordinates { (100, 44.8) (200,72.8) (300,82.0)};
            \addplot[dotted,color=blue] coordinates {(300,82.0) (400,86.3) (500,88.3) (600,89.9) (700,91.0) (800,91.5)}; 
        \end{axis}
    \end{scope}
    \node at (5.15, 2.5) {evaluation};

    \draw[-Latex,out=10,in=170] (dnn) to node[midway,below] {testing} (perfplot);

    % \draw[fill=gray!20, draw=none] (6.3, -1.5) rectangle (9.6, 2.4);

    % \begin{scope}[local bounding box=piechart,
    %     shift={(8, 0)},
    %     scale=0.45
    %     ]
    %     \pie[color={blue!30, purple!30},
    %         explode=0.1,
    %         hide number,
    %         rotate=-30
    %         % text=legend
    %     ]
    %     {
    %         96.7/,
    %         3.3/
    %     }
    % \end{scope}
    % \node at (8, 2) {compute time};
\end{tikzpicture}

% \end{document}

%% file: figures/diagram.tex
% \documentclass{standalone}
% \usepackage{tikz}
% \usepackage{graphicx}
% \usepackage{transparent}
% \usepackage{eso-pic}
% \usepackage{pgfplots}
% \pgfplotsset{compat=newest}

% \begin{document}
\begin{tikzpicture}
    \begin{scope}[cm={1,0,0.5,0.6,(0,0)}] % Sets the coordinate trafo matrix entries.
        \node[transform shape,opacity=0.8] at (1.5,-1) {\includegraphics[width=2cm]{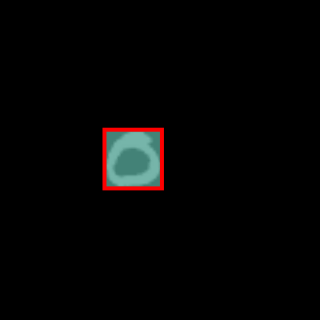}};
        \node[transform shape,opacity=0.8] at (1,0) {\includegraphics[width=2cm]{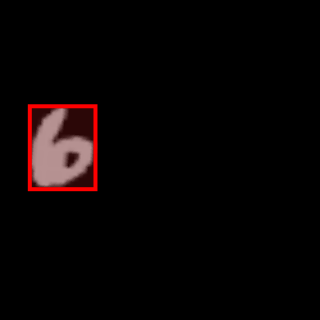}};
        \node[transform shape,opacity=0.8] at (0.5,1) {\includegraphics[width=2cm]{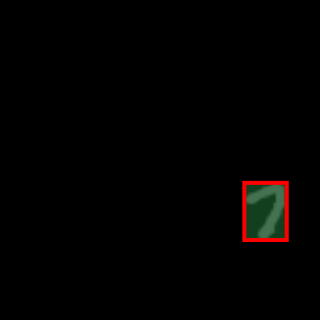}};
        \node[transform shape,opacity=0.8] at (0,2) {\includegraphics[width=2cm]{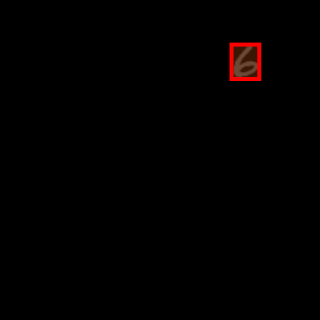}};
        \node[transform shape] at (2.5,-3) {\includegraphics[width=2.5cm]{figures/000000000025.jpg}};
    \end{scope}
    \node at (6, 0) {\includegraphics[width=2.5cm]{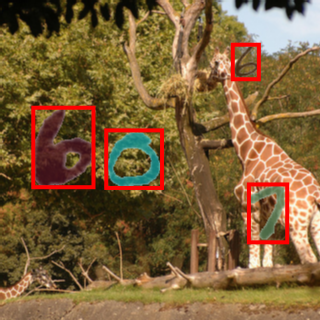}};
    \draw [-stealth] (2.7, 0) -- (4.5, 0);
    \node at (-2.4, -2) {\small COCO background};
    \node at (-2.4, 1.5) {\small MNIST numbers:};
    \node at (-2.4, 1.1) {\small transformed and colorized};

    \def\arcbegin{0}
    \def\arcending{280}
\begin{scope}[xshift=-3.7cm,yshift=-1.5cm,scale=0.4]
\begin{axis}[
    view={19}{30},
    axis lines=center,
    axis on top,
    domain=0:1,
    y domain=\arcbegin:\arcending,
    xmin=-1.5, xmax=1.5,
    ymin=-1.5, ymax=1.5,
    zmin=0.0, zmax = 1.2,
    hide axis,
    samples = 20,
    data cs=polar,
    mesh/color input=explicit mathparse,
    shader=interp,
    at={(-2, 0)}]
% cone:
\addplot3 [
    surf,
    variable=\u,
    variable y=\v,
    point meta={symbolic={Hsb=v,u,u}}] 
    (v,1,u);
% top plane:
\addplot3 [
    surf,
    samples = 50,
    variable=\u,
    variable y=\v,
    point meta={symbolic={Hsb=v,u,1}}] 
    (v,u,1);
% slice plane
\addplot3 [
    surf,
    variable=\u,
    y domain = 0:1,
    variable y=\w,
    point meta={symbolic={Hsb=\arcbegin,u,z}}]
    (\arcbegin,u,w);
\addplot3 [
    surf,
    variable=\u,
    y domain = 0:1,
    variable y=\w,
        point meta={symbolic={Hsb=\arcending,u,z}}]
    (\arcending,u,w);
% border
\addplot3[
    line width=0.3pt]
    coordinates {(0,0,0) (\arcbegin, 1, 0) (\arcbegin,1,1) (0,0,1) ({(\arcending)},1,1) (\arcending, 1, 0) (0,0,0) };
% border top
\draw[
    line width = 0.3pt]
    (axis cs: {cos(\arcbegin)}, {sin(\arcbegin)},1) arc (\arcbegin:\arcending:1);
% arc
\draw[
    -,
    line width = 1pt]
    (axis cs: {1.5*cos(\arcbegin+20)}, {1.5*sin(\arcbegin+20)},0.5) arc ({\arcbegin}:{71.3}:1.5);
\draw[
    ->,
    line width = 1pt]
    (axis cs: {1.5*cos(\arcending-130)}, {1.5*sin(\arcending-130)},0.5) arc ({\arcending-130}:{\arcending}:1.5);
% x and z axis
\addplot3[
    <->,
    line width=0.8pt]
    coordinates {(\arcbegin,1.5,0) (0,0,0) (0,0,1.5)};
% annotations
\node at (axis cs:1.3,0,0) [anchor=north east] {\Large S};
\node at (axis cs:-0.1,0,1.6) [anchor= north east] {\Large V};
\node at (axis cs:-1.65,0.0,0.4) [anchor=east] {\Large H};

\end{axis}
\end{scope}

    % \node at (-5, 0) {\small $(H, S, V) \sim \mathcal{U}([0, 1] \times [0.05, 1] \times [0.1, 1])$};

\end{tikzpicture}
% \end{document}

%% file: tables/num_images_queried.tex
\begin{tabular}{cl|cccc|cccc}
\toprule
{} & {} & \multicolumn{4}{c}{\# queried images} & \multicolumn{4}{c}{\# queried bounding boxes} \\
{} & {} & MNIST-Det & EMNIST-Det & Pascal VOC & BDD100k & MNIST-Det & EMNIST-Det & Pascal VOC & BDD100k \\
\midrule
\parbox[t]{0.05\linewidth}{\multirow{5}{*}{\rotatebox[origin=c]{90}{YOLOv3}}} 
& Random &  327.9 &       595.6 &   2236.8 &   5871.2 &      1079.1 &       1825.3 &    5344.2 &  116362.1 \\
& Entropy &      \textbf{245.5} &       \textbf{398.8} &   \underline{1732.8} &   5389.3 &      \textbf{1004.9} &       \textbf{1583.0} &    \textbf{4695.4} &  110694.9 \\
& Prob.~Margin &      256.2 &       429.0 &   1858.5 &   \textbf{4895.2} &      \underline{1013.7} &       1617.1 &    \underline{4787.6} &  \textbf{100376.3} \\
& MC Dropout &      256.3 &       416.2 &   \textbf{1679.4} &   \underline{5200.5} &      1115.3 &       1671.6 &    4875.1 &  \underline{110427.6} \\
& Mutual Inf. &      \underline{249.8} &       \underline{399.5} &   1884.2 &   5912.9 &      1061.9 &       \underline{1602.7} &    5527.0 &  125050.1 \\
\midrule 
\parbox[t]{0.05\linewidth}{\multirow{5}{*}{\rotatebox[origin=c]{90}{Faster R-CNN}}} 
& Random &       450.0 &        843.4 &    1293.7 &    6434.3 &       2140.0 &        2891.7 &     3125.2 &   129219.0 \\
& Entropy &       \textbf{384.5} &        \textbf{561.6} &    \textbf{1030.6} &    \underline{5916.7} &       \textbf{1608.4} &        \textbf{2156.4} &     \textbf{2707.0} &   \underline{123008.6} \\
& Prob.~Margin &       408.7 &        626.2 &    \underline{1036.5} &    \textbf{5761.6} &       \underline{1622.9} &        2285.1 &     \underline{2711.6} &   \textbf{117889.3} \\
& MC Dropout &       \underline{390.5} &        647.4 &    1127.5 &    6296.4 &       1818.1 &        2773.8 &     3624.7 &   130533.8 \\
& Mutual Inf. &       395.3 &        \underline{572.6} &    1080.2 &    6385.7 &       1695.6 &        \underline{2235.3} & 3026.5 &   132855.7 \\
\bottomrule
\end{tabular}

%% file: figures/timing_plot.tex
% \documentclass[border=0.2cm]{standalone}
 
% % Bar chart drawing library 
% \usepackage{pgfplots} 
 
% \begin{document}
 
\begin{tikzpicture}
 
\begin{axis} [ybar,
    ymode=log,
    log origin=infty,
    ylabel=time in hours,
    ymin=0,
    ymax=20,
    ymajorgrids=true,
    ytick={1, 10},
    yticklabels={1, 10},
    bar width = 10pt,
    enlarge x limits = {abs=0.5},
    xtick = {1, 2, 3},
    xticklabels = {EMNIST-Det, Pascal VOC, BDD100k},
    legend style={at={(0.05,0.95)},
        anchor=north west,
        draw=none,
        fill=none,
        nodes={scale=0.9, transform shape},
        legend cell align={left}
        }
    % x tick label style = {rotate=15, anchor=north}
]

% YOLOv3
\addplot coordinates {(1,0.6200665054553085) (2,8.906666667) (3,11.4262272919)};
% \addplot coordinates {(1.19,0.6200665054553085) (2.1,0.8793364878) (3.19,11.4262272919)};
% \addplot coordinates {(2.1,8.906666667)};
% RetinaNet
\addplot coordinates {(1,0.3611961772) (2,3.1586726786) (3,11.4841702413)};
% Faster R-CNN
\addplot coordinates {(1,0.8208678562) (2,4.147524456464582) (3,17.8681197843)};

\legend {YOLOv3, RetinaNet, Faster R-CNN}
 
\end{axis}

\node [rotate=90] at (0.7,1.6) {$0.62$};
\node [rotate=90] at (1.15,0.9) {$0.36$};
\node [rotate=90] at (1.55,2) {$0.82$};
% \node [rotate=90] at (2.8,2.1) {$0.88$};
\node [rotate=90] at (3,5.05) {$8.91$};
\node [rotate=90] at (3.4,3.7) {$3.16$};
\node [rotate=90] at (3.85,4.05) {$4.15$};
\node [rotate=90] at (5.3,4.5) {$11.43$};
\node [rotate=90] at (5.7,4.5) {$11.48$};
\node [rotate=90] at (6.13,5.05) {$17.87$};

\end{tikzpicture}
 
% \end{document}

%% file: tables/darknet_tables.tex
% \documentclass[]{standalone}

% \usepackage{booktabs}

% \begin{document}
    % \begin{table}
        \begin{tabular}{l c | c c | c c}\toprule
            & & \multicolumn{2}{c}{Darknet53} & \multicolumn{2}{c}{Darknet20} \\
            Type & Size & Blocks & Filters & Blocks & Filters \\\midrule
            Conv & $3 \times 3$ &  & $32$ &  & $32$ \\
            Conv & $3 \times 3 / 2$ &  & $64$ &  & $32$ \\\midrule

            Conv & $1 \times 1$ &  & $32$ &  & $32$ \\
            Conv & $3 \times 3$ & $1 \times$ & $64$ & $1 \times$ & $64$ \\
            Residual &  &  &  &  & \\\midrule
            
            Conv & $3 \times 3 / 2$ & & $128$ &  & $64$ \\\midrule

            Conv & $1 \times 1$ &  & $64$ &  & $32$ \\
            Conv & $3 \times 3$ & $2 \times$ & $128$ & $1 \times$ & $64$ \\
            Residual &  &  &  &  & \\\midrule

            Conv & $3 \times 3 / 2$ & & $256$ &  & $128$ \\\midrule

            Conv & $1 \times 1$ &  & $128$ &  & $64$ \\
            Conv & $3 \times 3$ & $8 \times$ & $256$ &  $1 \times$ & $128$ \\
            Residual &  &  &  &  & \\\midrule

            Conv & $3 \times 3 / 2$ & & $512$ &  & $256$ \\\midrule

            Conv & $1 \times 1$ &  & $256$ &  & $128$ \\
            Conv & $3 \times 3$ & $8 \times$ & $512$ & $2 \times$ & $256$ \\
            Residual &  &  &  &  & \\\midrule

            Conv & $3 \times 3 / 2$ & & $1024$ &  & $512$ \\\midrule

            Conv & $1 \times 1$ &  & $512$ &  & $256$ \\
            Conv & $3 \times 3$ & $4 \times$ & $1024$ & $2 \times$ & $512$ \\
            Residual &  &  &  &  & \\
            \bottomrule
        \end{tabular}
    % \end{table}
% \end{document}

%% file: tables/map_table.tex
\begin{tabular}{ll | CCCC}
\toprule
{} & &  \text{MNIST-Det} &  \text{EMNIST-Det} &  \text{Pascal VOC} &  \text{BDD100k} \\
\midrule
\parbox[t]{0.05\linewidth}{\multirow{5}{*}{\rotatebox[origin=c]{90}{YOLOv3}}} 
 & Random             &         89.28 &          88.95 &          72.10 &       38.22 \\
 & Entropy       &         \mathbf{91.53} &          \mathbf{91.75} &          \mathbf{73.08} &       \underline{39.28} \\
 & Prob. Margin  &         91.03 &          \underline{91.42} &          72.65 &       \mathbf{39.35} \\
 & MC Dropout    &         \underline{91.08} &          \underline{91.42} &          \underline{72.78} &       38.80 \\
 & Mutual Inf.     &         91.05 &          \underline{91.42} &          72.22 &       38.30 \\\midrule
\parbox[t]{0.05\linewidth}{\multirow{5}{*}{\rotatebox[origin=c]{90}{Faster R-CNN}}}
 & Random             &         83.20 &          83.92 &          74.67 &       47.02 \\
 & Entropy       &         \underline{85.27} &          \mathbf{87.15} &          \underline{75.10} &       \underline{48.30} \\
 & Prob. Margin  &         85.00 &          86.62 &          \mathbf{75.40} &       \mathbf{48.35} \\
 & MC Dropout    &         \mathbf{85.28} &          86.05 &          74.25 &       47.27 \\
 & Mutual Inf.     &         \underline{85.27} &          \underline{87.07} &          74.35 &       47.27 \\\midrule
\parbox[t]{0.05\linewidth}{\multirow{5}{*}{\rotatebox[origin=c]{90}{RetinaNet}}}
 & Random             &         82.95 &          83.70 &          69.58 &       43.20 \\
 & Entropy       &         \mathbf{85.35} &          \mathbf{86.15} &          \mathbf{71.80} &       \mathbf{43.97} \\
 & Prob. Margin  &         84.38 &          85.78 &          \underline{71.55} &       \underline{43.75} \\
 & MC Dropout    &         85.02 &          85.60 &          68.17 &       43.50 \\
 & Mutual Inf.     &         \underline{85.33} &          \underline{86.10} &          68.12 &       43.48 \\
\bottomrule
\end{tabular}

%% file: tables/auc_table.tex
\begin{tabular}{ll | CCCC|CCCC}
\toprule
& & \multicolumn{4}{c}{\# queried images} & \multicolumn{4}{c}{\# queried bounding boxes} \\
{} & & \text{MNIST-Det} & \text{EMNIST-Det} & \text{Pascal VOC} & \text{BDD100k} & \text{MNIST-Det} & \text{EMNIST-Det} & \text{Pascal VOC} & \text{BDD100k} \\
\midrule
\parbox[t]{0.05\linewidth}{\multirow{5}{*}{\rotatebox[origin=c]{90}{YOLOv3}}} &
Random             &         290.9 &          543.3 &         1645.1 &      1691.9 &        1503.8 &         2438.7 &         4274.3 &     37327.3 \\
 & Entropy       &         \mathbf{299.6} &          \underline{577.4} &         \mathbf{1685.2} &      \underline{1728.8} &        \mathbf{1519.0} &         \mathbf{2491.4} &         \mathbf{4324.6} &     \underline{37816.6} \\
 & Prob. Margin  &         298.1 &          571.9 &         1664.5 &      \mathbf{1733.6} &        \underline{1516.2} &         \underline{2485.3} &         \underline{4303.4} &     \mathbf{37931.1} \\
 & MC Dropout    &         298.3 &          571.8 &         \underline{1684.9} &      1727.9 &        1503.8 &         2465.5 &         4284.6 &     37550.7 \\
 & Mutual Inf.     &         \underline{299.1} &          \mathbf{578.0} &         1666.8 &      1716.1 &        1509.4 &         2482.6 &         4233.2 &     37514.5 \\
\midrule
\parbox[t]{0.05\linewidth}{\multirow{5}{*}{\rotatebox[origin=c]{90}{Faster R-CNN}}} &
Random             &         273.0 &          542.6 &         1207.9 &      2220.1 &        1448.6 &         2643.7 &         3121.5 &     47767.9 \\
 & Entropy       &         \mathbf{281.3} &         \mathbf{ 567.1} &         \mathbf{1237.6} &      \underline{2266.7} &        \mathbf{1465.2} &         \mathbf{2703.6} &         \mathbf{3168.8} &     \underline{48167.8} \\
 & Prob. Margin  &         279.7 &          561.3 &         \underline{1236.4} &      \mathbf{2274.7} &        \mathbf{1465.2} &         \underline{2692.5} &         \underline{3168.3} &     \mathbf{48449.2} \\
 & MC Dropout    &         \underline{280.5} &          557.0 &         1227.0 &      2247.3 &        1451.4 &         2619.7 &         3052.8 &     47718.9 \\
 & Mutual Inf.     &         280.2 &          \underline{566.6} &         1230.5 &      2247.8 &        \underline{1457.5} &         2697.4 &         3123.8 &     47714.8 \\
\midrule
\parbox[t]{0.05\linewidth}{\multirow{5}{*}{\rotatebox[origin=c]{90}{RetinaNet}}} &
Random             &         270.4 &          677.5 &         1743.6 &      1380.5 &        1438.2 &         2896.9 &         4614.9 &     33141.5 \\
 & Entropy       &         \mathbf{279.1} &          \mathbf{701.9} &         \mathbf{1819.7} &      \mathbf{1421.6} &        \underline{1446.2} &         \underline{2932.4} &         \underline{4714.6} &     33431.4 \\
 & Prob. Margin  &         276.8 &          696.5 &         \underline{1802.8} &      \underline{1414.7} &        1445.6 &         2928.4 &         \mathbf{4722.5} &     33339.5 \\
 & MC Dropout    &         \underline{279.0} &          696.3 &         1730.6 &      1406.2 &        1445.6 &         2915.8 &         4535.9 &     \underline{33452.9} \\
 & Mutual Inf.     &         278.5 &          \underline{699.3} &         1734.7 &      1400.8 &        \mathbf{1449.1} &         \mathbf{2940.7} &         4543.9 &     \mathbf{33495.2} \\
\bottomrule
\end{tabular}

%% file: tables/retinanet_table.tex
\begin{tabular}{ll| cccc |cccc}
\toprule
& & \multicolumn{4}{c}{\# queried images} & \multicolumn{4}{c}{\# queried bounding boxes} \\
{} & & MNIST-Det & EMNIST-Det & Pascal VOC & BDD100k & MNIST-Det & EMNIST-Det & Pascal VOC & BDD100k \\
\midrule
\parbox[t]{0.05\linewidth}{\multirow{5}{*}{\rotatebox[origin=c]{90}{RetinaNet}}}  & Random             &         $390.3$ &          $950.4$ &         $2555.4$ &      $3616.2$ &        $1283.8$ &         $2957.7$ &         $6220.0$ &     $69842.0$ \\
 & Entropy (sum)      &         $\mathbf{288.6}$ &          $\mathbf{687.7}$ &        $\mathbf{1961.2}$ &      $\mathbf{2866.5}$ &        $1292.0$ &         $\underline{2708.6}$ &         $\mathbf{5421.6}$ &     $64939.7$ \\
 & Prob.~ Margin (sum) &         $310.8$ &          $733.9$ &         $\underline{2087.3}$ &      $\underline{2901.5}$ &        $\underline{1277.5}$ &         $2721.5$ &         $\underline{5445.6}$ &     $64794.9$ \\
 & MC Dropout (sum)   &         $293.3$ &          $749.6$ &         $2745.3$ &      $3027.7$ &        $1317.4$ &         $2926.4$ &         $7047.9$ &     $\underline{62395.5}$ \\
 & Mutual Inf. (sum)    &         $\underline{289.6}$ &          $\underline{719.0}$ &         $2881.9$ &      $3124.9$ &        $\mathbf{1248.0}$ &         $\mathbf{2677.4}$ &         $7389.0$&     $\mathbf{61712.9}$ \\
\bottomrule
\end{tabular}